\documentclass[letterpaper]{article} 
\usepackage{aaai2026}  
\usepackage{times}  
\usepackage{helvet}  
\usepackage{courier}  
\usepackage[hyphens]{url}  
\usepackage{graphicx} 
\usepackage{natbib}  
\usepackage{caption} 
\usepackage{algorithm}
\usepackage{algorithmic}
\usepackage{amsmath}
\usepackage{amsfonts}
\usepackage{booktabs}
\usepackage{multirow}
\usepackage{colortbl}
\usepackage[table]{xcolor}
\usepackage{makecell}
\usepackage{amsmath}
\usepackage{tcolorbox}
\usepackage{etoolbox}
\usepackage{needspace}
\usepackage{mdframed}
\usepackage{newfloat}
\usepackage{listings}
\DeclareCaptionStyle{ruled}{labelfont=normalfont,labelsep=colon,strut=off} 
\floatstyle{ruled}
\newfloat{listing}{tb}{lst}{}
\floatname{listing}{Listing}

\newcounter{appsection}
\newcounter{appsubsection}[appsection]

\renewcommand\theappsection{\Alph{appsection}}
\renewcommand\theappsubsection{\theappsection.\arabic{appsubsection}}

\newcommand{\appsection}[1]{%
  \refstepcounter{appsection}%
  \section*{\fontsize{10}{12}\selectfont \theappsection. \quad #1} 
  \addcontentsline{toc}{section}{\theappsection. #1}%
}

\newcommand{\appsubsection}[1]{%
  \refstepcounter{appsubsection}%
  \subsection*{\fontsize{10}{12}\selectfont \theappsubsection. \quad #1} 
  \addcontentsline{toc}{subsection}{\theappsubsection. #1}%
}

\lstdefinelanguage{JSON}{
  morestring=[b]",
  morecomment=[l][\color{gray}]{\#},
  stringstyle=\ttfamily\color{purple},
  keywordstyle=\color{blue},
  commentstyle=\color{gray},
  morekeywords={true,false,null},
}

\newmdenv[
  linecolor=black,
  linewidth=0.8pt,
  topline=true,
  bottomline=true,
  rightline=true,
  leftline=true,
  backgroundcolor=white,
  skipabove=\baselineskip,
  skipbelow=\baselineskip,
  innerleftmargin=10pt,
  innerrightmargin=10pt,
  innertopmargin=10pt,
  innerbottommargin=10pt,
  roundcorner=5pt,
  splittopskip=\topskip,
  splitbottomskip=\topskip,
]{mybox}

\title{ManiLong-Shot: Interaction-Aware One-Shot Imitation Learning for Long-Horizon Manipulation}
\author{
    Zixuan Chen\textsuperscript{\rm 1}\thanks{This work was partially conducted during Zixuan Chen's visit to National University of Singapore.},
    Chongkai Gao\textsuperscript{\rm 2},
    Lin Shao\textsuperscript{\rm 2}\thanks{Corresponding author: Lin Shao, Jing Huo},
    Jieqi Shi\textsuperscript{\rm 1,3},
    Jing Huo\textsuperscript{\rm 1}\footnotemark[2],
    Yang Gao\textsuperscript{\rm 4,1,3}
}
\affiliations{
    \textsuperscript{\rm 1}State Key Laboratory for Novel Software Technology, Nanjing University, Nanjing, China\\
    \textsuperscript{\rm 2}School of Computing, National University of Singapore, Singapore\\
    \textsuperscript{\rm 3}School of Intelligence Science and Technology, Nanjing University, Suzhou, China\\
    \textsuperscript{\rm 4}School of Network Security and Information Technology, YiLi Normal University, Xinjiang, China\\
    chenzx@nju.edu.cn, linshao@nus.edu.sg, huojing@nju.edu.cn
}

\usepackage{bibentry}

\begin{document}

\maketitle

\begin{abstract}
One-shot imitation learning (OSIL) offers a promising way to teach robots new skills without large-scale data collection. However, current OSIL methods are primarily limited to short-horizon tasks, thus limiting their applicability to complex, long-horizon manipulations. To address this limitation, we propose ManiLong-Shot, a novel framework that enables effective OSIL for long-horizon prehensile manipulation tasks.
ManiLong-Shot structures long-horizon tasks around physical interaction events, reframing the problem as sequencing interaction-aware primitives instead of directly imitating continuous trajectories. This primitive decomposition can be driven by high-level reasoning from a vision-language model (VLM) or by rule-based heuristics derived from robot state changes. For each primitive, ManiLong-Shot predicts invariant regions critical to the interaction, establishes correspondences between the demonstration and the current observation, and computes the target end-effector pose, enabling effective task execution.
Extensive simulation experiments show that ManiLong-Shot, trained on only 10 short-horizon tasks, generalizes to 20 unseen long-horizon tasks across three difficulty levels via one-shot imitation, achieving a \textbf{22.8\%} relative improvement over the SOTA. Additionally, real-robot experiments validate ManiLong-Shot’s ability to robustly execute three long-horizon manipulation tasks via OSIL, confirming its practical applicability. 
\end{abstract}

\begin{links}
    \link{Website}{https://sites.google.com/view/manilong-shot}
\end{links}

\section{Introduction}

For robots to effectively integrate into daily life, they must rapidly learn and execute diverse long-horizon manipulation tasks. Daily chores, such as ``setting the table'' or ``tidying the kitchen'', exemplify these challenges, as they typically involve sequential interactions with multiple objects and comprise a series of sub-tasks. One-shot imitation learning (OSIL)~\cite{duan2017one,finn2017one,valassakis2022demonstrate} is a key technology for achieving this goal, allowing robots to acquire new skills from a single demonstration while avoiding costly retraining.
However, despite its promise, existing OSIL methods often struggle to scale to long-horizon tasks. Many approaches are limited to short-horizon skills~\cite{zhang2024oneshot, vosylius2024instant}, require new tasks to be slight variations of training tasks~\cite{xu2022prompting}, or rely on known 3D object models~\cite{biza2023one}. These limitations significantly impede their application in complex, multi-stage, real-world manipulation scenarios.
We draw inspiration from human learning: when faced with an unseen task such as \textit{placing tableware}, a person naturally decomposes it into short-horizon primitives and infers the key interaction regions for each. For example: 1)~pick up a plate (rim); 2)~place the plate (target location on the table); 3)~pick up a fork (handle). Each primitive is bounded by physical interactions, such as contact or release, and imitation is achieved by replicating actions on these regions. This raises the question: \textit{Can a robot infer sub-task boundaries and critical interaction regions from an unannotated demonstration of an unseen long-horizon task, enabling efficient OSIL of its primitives?}

\begin{figure}[t!]
    \centering
    \includegraphics[width=\linewidth]{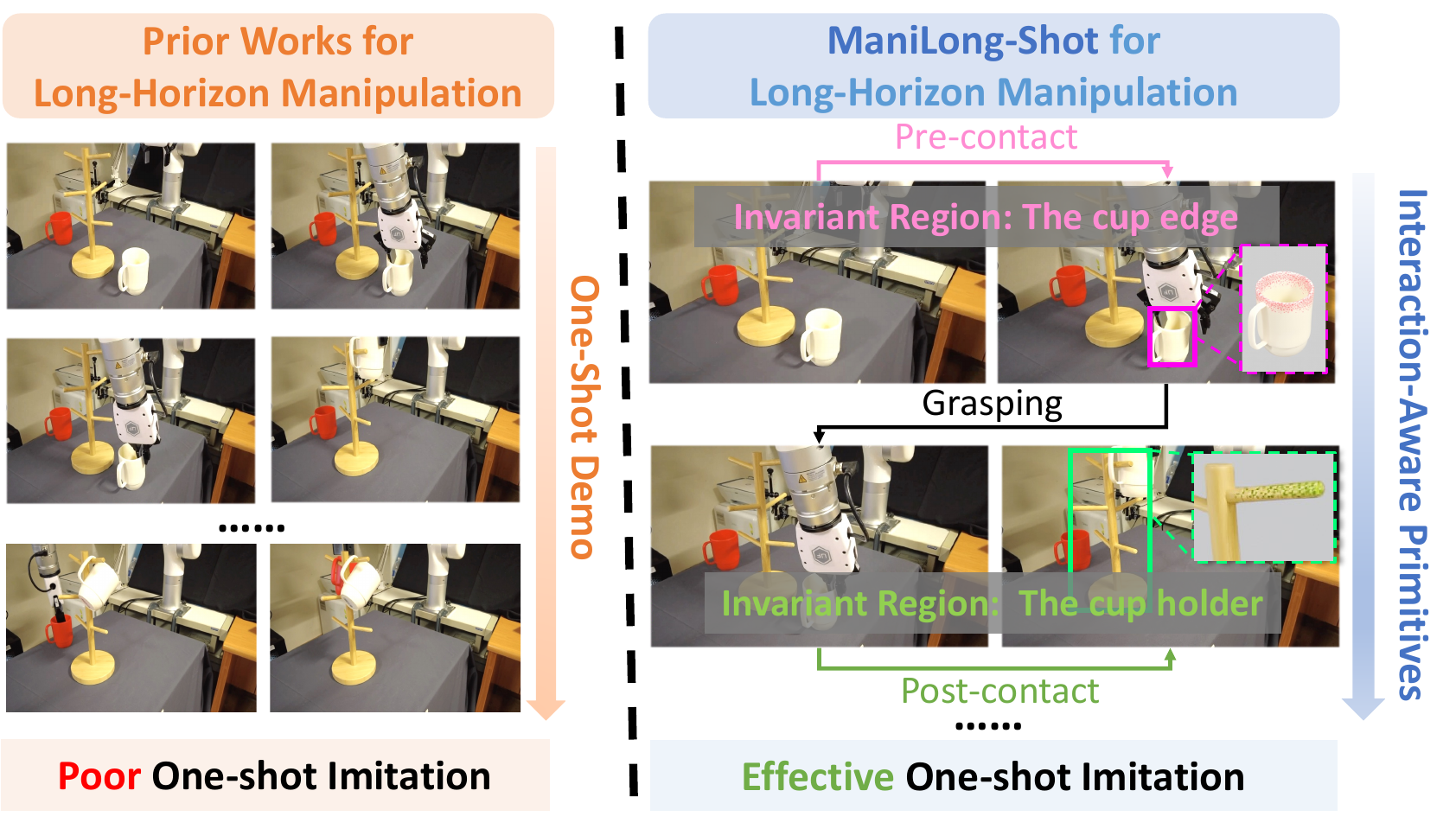}
    \caption{\small{We introduce ManiLong-Shot, a novel framework for effective OSIL in long-horizon prehensile manipulation.}} 
    \label{fig:teaser}
\end{figure}

To this end, we propose a novel interaction-aware OSIL framework, \textbf{ManiLong-Shot}, inspired by human strategies and cognitive patterns when performing OSIL for long-horizon manipulation. ManiLong-Shot structures long-horizon manipulation tasks into a sequence of discrete primitives, each corresponding to a distinct phase of physical interaction between the robot's end-effector and the environment, namely: \textit{pre-contact}, \textit{grasping}, and \textit{post-contact}, as illustrated in Fig.~\ref{fig:teaser}. 
The phase decomposition design builds upon prior work on sub-task decomposition for manipulation~\citep{james2022q,chen2024gravmad,chen2025deco}, and is founded on a key assumption: sub-task boundaries can be reliably identified through these interaction phases. Consequently, our framework primarily focuses on prehensile manipulation tasks, such as pick-and-place, and does not address purely non-prehensile behaviors.

ManiLong-Shot comprises three core modules:
\textbf{First}, it enables flexible task decomposition of long-horizon tasks from a single demonstration, utilizing either a simple heuristic based on gripper status and joint velocity or the semantic reasoning capabilities of a Vision-Language Model (VLM). Both strategies are implemented and systematically evaluated within our framework.
\textbf{Second}, after decomposing the one-shot demonstration into primitives, ManiLong-Shot focuses on robustly imitating each primitive. It employs an interaction-aware region prediction network that identifies crucial object surface regions for each interaction. For instance, in the cup-grasping task shown in Fig~\ref{fig:teaser}, the key action involves grasping the cup's edge—a region that consistently serves this function during the pre-contact phase, regardless of the cup’s pose. Such regions are termed \textit{invariant regions}~\cite{zhang2024oneshot}, defined as semantically and functionally stable interaction surfaces that generalize across diverse task scenes.
\textbf{Finally}, ManiLong-Shot introduces an interaction-aware region matching network that aligns the predicted invariant regions from the demonstration with their counterparts in the current scene during execution. This matching yields the target pose for the robot’s end-effector, enabling robust sequential execution of each primitive. Consequently, the robot can efficiently and reliably accomplish long-horizon tasks in a one-shot, fully end-to-end OSIL manner.

In summary, our contributions are as follows:
(1) We propose \textbf{ManiLong-Shot}, a novel OSIL framework designed for long-horizon prehensile manipulation tasks, which structures tasks into sequential interaction primitives and enables effective end-to-end OSIL execution through an interaction-aware pipeline.
(2) We introduce an interaction-aware task decomposition mechanism that utilizes VLMs or rule-based robot state transitions to derive interaction primitives from a single unannotated long-horizon demonstration. Each primitive undergoes a two-stage process to predict functionally invariant regions and perform region matching for accurate goal poses, ensuring robust task execution.
(3) We construct the \textbf{RLBench-Oneshot} benchmark by selecting 30 tasks from the manipulation benchmark RLBench~\citep{james2020rlbench} for systematic evaluation of our framework. Furthermore, we demonstrate the effectiveness of ManiLong-Shot’s OSIL capabilities on three real-world long-horizon manipulation tasks, validating its practical applicability.

\section{Related Work}
\paragraph{OSIL for Manipulation.}
One-shot imitation learning (OSIL) aims to learn from demonstrations of training tasks and generalize to novel tasks using only a single trajectory per new task, without additional training~\cite{duan2017one,finn2017one,yu2018one,biza2023one,zhang2024oneshot,bonardi2020learning}. It is a key technique for enabling generalization in manipulation tasks. Early work employs meta-learning to transfer knowledge across diverse robotic tasks~\cite{finn2017one,gao2023transferring,yu2018one}, or trains transformer-based policies using expert trajectories~\cite{mandi2022towards}. Wen et al.~\cite{wen2022you,wen2022catgrasp} adopt a ``pre-training and adaptation'' strategy to learn canonical representations, while others explore graph-based representations of task structure or scene geometry~\cite{zhang2024oneshot,kumar2023graph,huang2019neural}.
However, most existing approaches are limited to simple, short-horizon tasks. \citeauthor{wu2024one} (\citeyear{wu2024one}) extend OSIL to long-horizon extrinsic manipulation by composing short-horizon, goal-conditioned primitives. In contrast, our work eliminates the reliance on predefined primitive libraries and infers the interaction process directly from a single demonstration, providing a more practical and scalable solution for long-horizon prehensile manipulation.

\paragraph{Task Decomposition for Long-horizon Tasks.}

A common approach to long-horizon manipulation is the ``divide and conquer'' strategy, which breaks complex tasks into shorter, relatively independent sub-tasks. These sub-tasks can be manually defined~\cite{dalal2021accelerating,iim}, predefined by the environment~\cite{chen2025scar,zhu2021hierarchical}, or learned automatically~\cite{masson2016reinforcement,flip}. However, ensuring their reusability remains a challenge—changes in task goals or environments can render predefined boundaries ineffective, hindering generalization and robustness~\cite{wu2024one}. Recent studies propose decomposition methods based on physical interactions between the robot gripper and the environment, demonstrating strong robustness and transferability~\cite{chen2024gravmad,chen2025deco,chen2025robohiman}.
With the rise of vision-language models (VLMs), research increasingly explores their reasoning and generative capabilities for sub-task planning~\cite{huang2025foundation,ding2025lavira}. VLMs have been used to generate sub-goal sequences~\cite{myerspolicy,curtistrust,zhu2025adaptpnp}, synthesize executable manipulation code~\cite{liang2023code,huang2023voxposer,chen2025robohorizon}, and identify consistent keypoints across instances~\cite{fang2024keypoint}, enhancing the flexibility and generalization of task decomposition.
Our work builds on methods that leverage robot-object interactions~\cite{chen2024gravmad,chen2025deco}, proposing an interaction-aware framework for sub-task segmentation. This approach integrates heuristic changes in gripper state and VLMs to identify sub-task boundaries, providing a robust decomposition strategy for long-horizon prehensile manipulation tasks.

\begin{figure*}
    \centering
    \includegraphics[width=\textwidth]{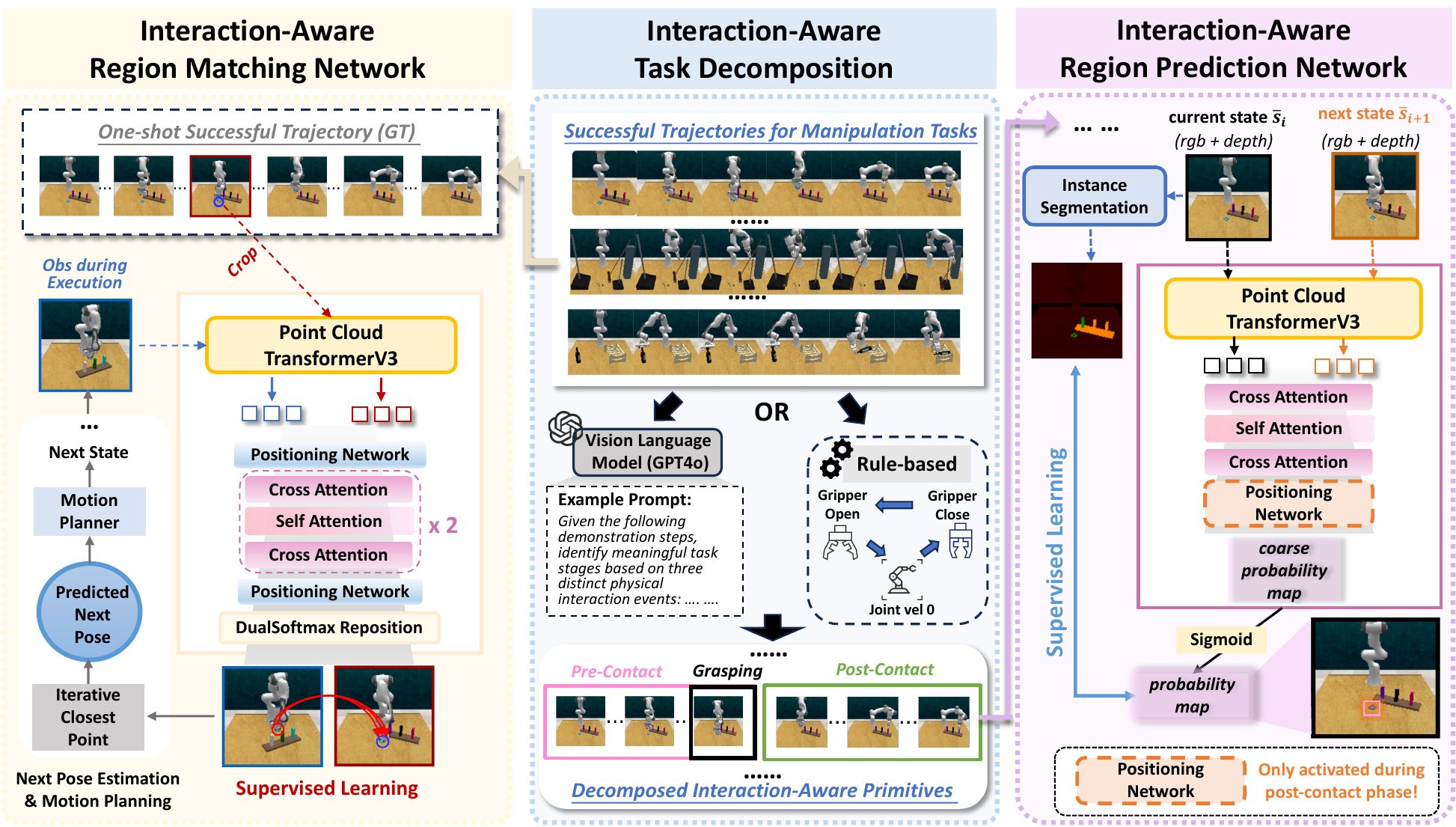}
    \caption{\small {\textbf{The Overall Training Pipeline of ManiLong-Shot.} Best viewed when zoomed in.}
    }
    \label{fig:training}
\end{figure*}

\paragraph{Invariant Correspondence for Manipulation.}
Establishing correspondences between seen and unseen scenarios~\citep{zhang2024tale} and incorporating invariance~\citep{brandstetter2021geometric,lyle2020benefits,riemann} are crucial for generalization in robotic systems. In robotic manipulation, prior work has utilized self-supervised learning to build dense visual correspondences~\citep{florence2018dense} or establish shape correspondences within object categories~\citep{wen2022transgrasp}, enabling grasp transfer across similar objects. Other approaches focus on learning viewpoint-invariant representations to facilitate action transfer from different perspectives~\citep{graf2023learning}. Recently, \citeauthor{zhang2024oneshot}(\citeyear{zhang2024oneshot}) proposed training neural networks to identify semantically invariant regions relevant to the robot’s end-effector, enabling one-shot action transfer. Our work builds on this by predicting and matching invariant regions across multiple phases of interaction-aware primitives, supporting effective one-shot generalization even in complex long-horizon manipulation tasks.

\section{Problem Formulation}
\label{formulation}
Our work addresses the problem of enabling a robot to achieve one-shot generalization to novel long-horizon prehensile manipulation tasks, given only a single unannotated successful demonstration. We focus on a class of representative tasks in which the robot operates a parallel-jaw gripper to grasp objects and perform pick-and-place interactions with multiple objects in a tabletop environment. Under this setting, we formally define the short-horizon and long-horizon manipulation tasks considered in this work as follows:
Based on the three physical interaction phases between the gripper and a single object—\textit{pre-contact}, \textit{grasping}, and \textit{post-contact}—a task is defined as \textit{short-horizon (SH)} if the gripper interacts with an object no more than once, i.e., the total number of such interactions is at most three. Otherwise, if the number of gripper-object interactions exceeds three, such as when manipulating multiple objects, the task is considered \textit{long-horizon (LH)}.
Formally, the manipulation tasks in this work are divided into two subsets: a SH set $\mathcal{T}^{\text{sh}}$, which provides abundant offline demonstrations for training, and a LH set $\mathcal{T}^{\text{lh}}$, which is unseen during training and requires one-shot generalization. During training, the robot learns from a dataset of demonstrations $\mathcal{D} = \{\tau^{(k)}_{\text{sh}} = \{(\overline{s}_t^{(k)}, \overline{a}_t^{(k)}, \overline{s}_{t+1}^{(k)})\}_{t=1}^{T_k}\}_{k=1}^K$, sampled from $\mathcal{T}^{\text{sh}}$, where each $\tau^{(k)}$ is a sequence of state-action pairs.
At test time, the robot is provided with a single successful demonstration for each novel task in $\mathcal{T}^{\text{lh}}$, denoted as $\tau^{\text{lh}} = \{(\hat{s}_t, \hat{a}_t, \hat{s}_{t+1})\}_{t=0}^{H}$, where $\hat{a}_t$ transitions the system from $\hat{s}_t$ to $\hat{s}_{t+1}$.
Given any state $s$ encountered during execution of novel LH task, the robot aims to predict an 18-dimensional action $a = (\mathbf{T}, \lambda)$ based on $\tau^{\text{lh}}$, where $\mathbf{T} \in \mathbb{R}^{4 \times 4}$ specifies the desired end-effector pose in $SE(3)$, and $\lambda \in \{0,1\}^2$ encodes the gripper command and a collision flag.
Low-level motion execution for $\mathbf{T}$ is handled by standard motion planners. The robot iteratively predicts actions and plans motions until the whole task is completed.

\section{Method}

\paragraph{Framework Overview.}
We propose \textbf{ManiLong-Shot}, an interaction-aware one-shot imitation learning (OSIL) framework for long-horizon prehensile manipulation. It consists of three modules: (1) \textbf{Interaction-aware Task Decomposition}, which decomposes a demonstration into primitives based on three physical interaction events; (2) \textbf{Interaction-aware Region Prediction Network}, which predicts functionally and semantically invariant regions for each sub-task; and (3) \textbf{Interaction-aware Region Matching Network}, which aligns these regions with novel scene observations to ensure spatial correspondence. After matching, the system estimates and executes the next end-effector pose via motion planning, enabling sequential task execution. ManiLong-Shot achieves effective one-shot generalization to complex, long-horizon manipulation tasks. Fig.~\ref{fig:training} illustrates the training pipeline of ManiLong-Shot.

\paragraph{Interaction-aware Task Decomposition.}
\label{sec:decomposition}
\begin{figure}
    \centering
    \includegraphics[width=\linewidth]{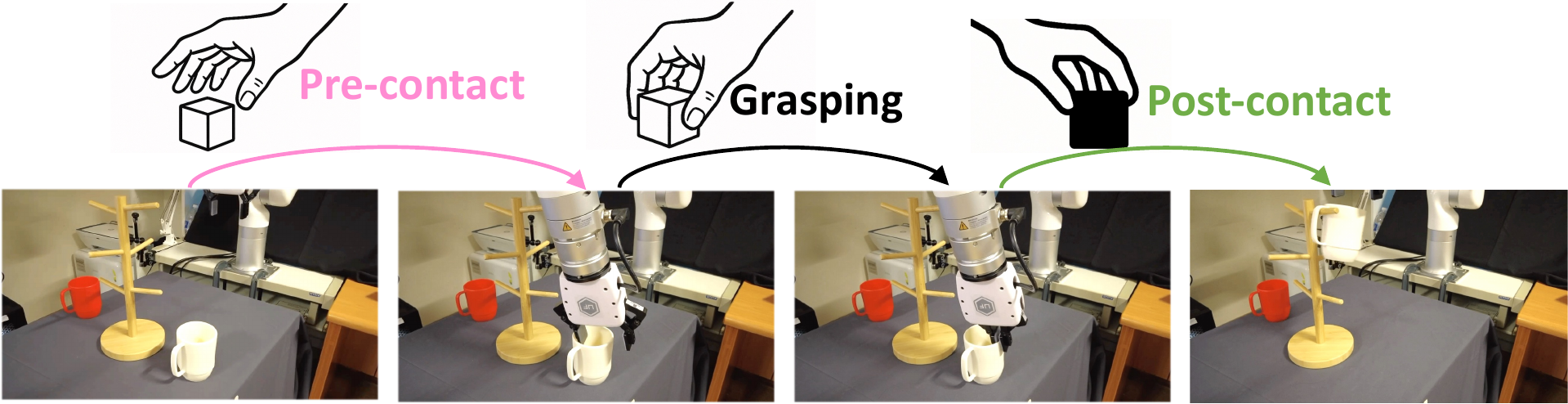}
    \caption{\small Visualization of the three physical interaction phases.}
    \label{fig:decom_real}
\end{figure}

This process organizes successful demonstration trajectories into a sequence of interaction-aware primitives based on physical interaction phases. We focus on prehensile manipulation tasks with a gripper, drawing inspiration from human strategies in OSIL for long-horizon tasks. Following existing sub-task decomposition schemes~\cite{james2022q, chen2024gravmad, chen2025deco}, we categorize robot-object interactions into three phases: \textit{pre-contact}, \textit{grasping}, and \textit{post-contact}. Fig.~\ref{fig:decom_real} illustrates these phases in the \textit{Place 1 Cup on Cupholder} task with a real robot, highlighting their correspondence with human cognitive understanding of manipulation.
Demonstration trajectories can be decomposed in two ways. One approach uses tailored prompts for vision-language models (VLMs) and structured representations of the trajectory to guide the automatic identification of interaction phases. In ManiLong-Shot, we employ GPT-4o~\cite{hurst2024gpt} as the underlying VLM. Alternatively, a rule-based method analyzes joint velocity and gripper state changes to infer phase boundaries throughout the trajectory.
The \textit{pre-contact} phase includes frames where the gripper remains open while approaching the object, ending when joint velocity drops to zero before closure, indicating precise alignment. The \textit{grasping} phase captures the transition from an open to a closed gripper securing the object. The \textit{post-contact} phase follows a successful grasp and involves transitioning from a closed to an open gripper while placing the object or interacting with others.
Short-horizon tasks typically involve a single cycle of these phases, while long-horizon tasks repeat them until completion. Both decomposition methods are interchangeable in our framework: the rule-based method offers stability, while the VLM-based approach enables learning of semantically aware interaction patterns.

\paragraph{Interaction-aware Region Prediction Network.}
\label{sec:region_predict}
After extracting interaction-aware subtask primitives from demonstrations, ManiLong-Shot predicts functionally \textbf{interaction-invariant} regions across interaction stages. To achieve this, we train an Interaction-aware Region Prediction Network (Fig.~\ref{fig:training}(right)) using only successful short-horizon (SH) demonstrations, highlighting the system’s data efficiency. The network builds on the \textit{invariant region} concept from~\citeauthor{zhang2024oneshot}~(\citeyear{zhang2024oneshot}), where a 3D geometric subset is invariant if it preserves structure equivariant to $SE(3)$ transformations in states with the same optimal policy. For example, in the ``insert onto square peg'' task (Fig.~\ref{fig:training}), the ``square ring'' in the \textit{pre-contact} phase consistently aligns with the grasp surface's principal axis, despite pose variations, demonstrating strong invariant properties.

Training the Interaction-aware Region Prediction Network utilizes demonstrations from the SH task set $\mathcal{T}^{\text{sh}}$. For each task $\mathcal{T}_i^{\text{sh}}$, one demonstration is divided into three segments: $\tau_{pre}^{sh}$, $\tau_{grasp}^{sh}$, and $\tau_{post}^{sh}$. For each pair of consecutive states $\{\overline{s}_i, \overline{s}_{i+1}\}$, dense point clouds are generated from RGB-D observations and processed by the Point Cloud Transformer V3 (PTV3)~\cite{wu2024point}. This involves progressive downsampling with cross-attention across scenes ($\overline{s}_i$ and $\overline{s}_{i+1}$) and self-attention within each scene, enhancing inter-state correspondence and intra-state feature representation.
The Positioning Network is activated only during the \textit{post-contact} phase, where the robot aligns the grasped object with a target region using attention mechanisms. Since the invariant regions in the \textit{pre-contact} and \textit{grasping} phases are functionally similar, we train their prediction networks jointly. The network generates a coarse saliency map from learned spatial priors, refines it through point-wise sigmoid parameterization, and outputs an interaction probability distribution to activate the relevant region $\mathcal{I}(\overline{s}_i)$, yielding the predicted invariant region (square ring in Fig.~\ref{fig:training}). The pipeline is trained with supervised learning using ground-truth instance masks from simulation for accurate supervision.

\paragraph{Interaction-aware Region Matching Network.}
\label{sec:region_matching}
Another key module of ManiLong-Shot is the Interaction-aware Region Matching Network. While the Interaction-aware Region Prediction Network identifies invariant regions across interaction primitives in demonstrations, this module enables the robot to match these regions to its current execution state for action reproduction without additional training. The network architecture builds on prior work~\cite{zhang2024oneshot}, as illustrated in Fig.~\ref{fig:training}(left). It is trained on multiple SH task demonstrations without explicit task decomposition, with each state annotated with GT invariant regions derived from instance segmentation masks and RGB-D data.
For a given SH task, consider a rollout trajectory $\tau^{sh}_j$ that differs from the demonstration $\tau^{sh}_i$. For each state $\overline{s}_j$ in $\tau^{sh}_j$, a state routing network~\cite{zhang2024oneshot} selects the most similar state $\overline{s}_i$ in $\tau^{sh}_i$ based on feature similarity. The invariant region $I(\overline{s}_i)$ is cropped from its point cloud and fused with the full-scene point cloud of $\overline{s}_j$, jointly downsampled by the PTV3 model. The Positioning Network applies an initial attention step followed by a two-stage cross–self–cross attention module to enhance spatial and geometric feature alignment between the two clouds, producing point-wise features over $\overline{s}_j$.
Based on these features, a dual-softmax matching algorithm~\cite{li2022lepard} computes a correspondence matrix $\mathbf{C}$ between the scene cloud of $\overline{s}_j$ and the invariant region cloud of $\overline{s}_i$. The network is trained using GT correspondence matrices and object position labels between $\tau^{sh}_i$ and $\tau^{sh}_j$ at each timestep.

After obtaining the correspondence matrix $\mathbf{C}$ between the invariant region $\mathcal{I}(\overline{s}_i)$ of the demonstrated state and the current execution state $\overline{s}_j$, we apply a correspondence-based pose regression algorithm~\cite{zhang2024oneshot} to estimate the robot's action pose $\mathbf{T}_j$ at $\overline{s}_j$. The optimization objective is:
$\mathbf{T}_j = \arg\min_{\mathbf{T} \in \mathrm{SE}(3)} 
\left\| \mathbf{T} \, \mathbf{T}_i^{-1} \, P_{\mathcal{I}(\overline{s}_i)} \, \mathbf{C} - P_{\overline{s}_j} \right\|,$
where $\mathbf{T}_i$ denotes the demonstrated action pose at $\overline{s}_i$, $P_{\mathcal{I}(\overline{s}_i)}$ and $P_{\overline{s}_j}$ represent the point clouds of the invariant region in $\overline{s}_i$ and the full scene in $\overline{s}_j$, respectively. 
Once the next pose $\mathbf{T}_j$ is determined, we use the RRT-Connect algorithm~\cite{lavalle2001randomized} for path planning, computing a collision-free trajectory from the current state to the target state and determining the next state observation after executing the predicted pose. This observation is then fed into the state routing network for comparison with frames from the demonstration trajectory, initiating the next round of invariant region matching. This iterative process continues until the entire task is completed.
More training details are provided in Appendix~\ref{app:training_archi}.

\paragraph{Evaluation of ManiLong-Shot}
These three key modules together form the ManiLong-Shot framework. When evaluating unseen long-horizon (LH) tasks, ManiLong-Shot first uses the Interaction-Aware Task Decomposition Module to break down a successful demonstration into a sequence of interaction-aware primitives. For each interaction phase, the Interaction-Aware Region Prediction Network predicts the corresponding invariant region.
During execution, the system captures real-time observations and employs the Interaction-Aware Region Matching Network to align the predicted regions with the current observations. Based on the correspondence matrix, it performs pose prediction and motion planning. This iterative process continuously refines poses and actions through region matching until the entire long-horizon task is completed. 
Evaluation details are provided in Appendix~\ref{app:eval}.

\begin{table*}[t]
\centering
\scriptsize
\setlength{\tabcolsep}{2.0pt}
\renewcommand{\arraystretch}{0.9}
\begin{tabular}{lcccccccccccc} 
\toprule
& Avg. & Avg. & Turn & Open & Put Groceries & Place Shape & Put Money & Close & Place & Light & Insert & Meat \\
Models & Success (\%) $\uparrow$ & Rank $\downarrow$ & Tap & Drawer & in Cupboard & in Shape Sorter & in Safe & Jar & Wine & Bulb In & Peg & off Grill \\
\midrule
RVT2 & 79.2 & 3.1 & 99.0 {\scriptsize± 1.5} & 74.0 {\scriptsize± 4.5} & 66.0 {\scriptsize± 5.6} & 35.0 {\scriptsize± 6.8} & 96.0 {\scriptsize± 2.0} & \textbf{100.0 {\scriptsize± 0.0}} & 95.0 {\scriptsize± 2.4} & 88.0 {\scriptsize± 3.2} & 40.0 {\scriptsize± 6.2} & 99.0 {\scriptsize± 1.2} \\
3DDA & 85.0 & 2.9 & \textbf{99.2} {\scriptsize± 1.6} & 89.6 {\scriptsize± 4.1} & \textbf{85.6} {\scriptsize± 4.1} & 44.0 {\scriptsize± 4.4} & 97.6 {\scriptsize± 2.0} & 96.0 {\scriptsize± 3.6} & 93.6 {\scriptsize± 4.8} & 82.4 {\scriptsize± 2.0} & 65.6 {\scriptsize± 4.1} & 96.8 {\scriptsize± 1.6} \\
ARP & 86.6 & 2.7 & 100.0 {\scriptsize± 0.0} & 92.8 {\scriptsize± 2.2} & 69.6 {\scriptsize± 5.2} & 46.4 {\scriptsize± 6.3} & 94.4 {\scriptsize± 2.5} & 97.6 {\scriptsize± 1.5} & 92.0 {\scriptsize± 2.1} & 94.4 {\scriptsize± 1.8} & \textbf{78.4} {\scriptsize± 3.9} & 96.0 {\scriptsize± 1.7} \\
IMOP & 65.24 & 3.9 & 51.2 {\scriptsize± 5.9} & 100.0 {\scriptsize± 0.0} & 46.4 {\scriptsize± 6.5} & 37.6 {\scriptsize± 7.0} & 96.0 {\scriptsize± 2.4} & 39.2 {\scriptsize± 6.7} & 96.0 {\scriptsize± 2.3} & 82.0 {\scriptsize± 4.1} & 12.0 {\scriptsize± 9.5} & 92.0 {\scriptsize± 2.7} \\
\rowcolor[HTML]{EFEFEF}
\textbf{ManiLong-Shot} & \textbf{90.4} \textcolor{black}{(\textbf{3.8\% $\uparrow$})} & \textbf{1.9} & 95.4 {\scriptsize± 2.1} & \textbf{100.0} {\scriptsize± 0.0} & 82.0 {\scriptsize± 4.2} & \textbf{48.0} {\scriptsize± 6.3} & \textbf{100.0} {\scriptsize± 0.0} & 96.8 {\scriptsize± 1.9} & \textbf{100.0} {\scriptsize± 0.0} & \textbf{88.0} {\scriptsize± 2.8} & 44.0 {\scriptsize± 6.5} & \textbf{100.0} {\scriptsize± 0.0} \\
\bottomrule
\end{tabular}
\caption{\small{\textbf{Performance on 10 Training SH Tasks.} We report the mean and standard deviation of success rates over 5 random seeds for each task, along with the average success rate and average rank across all tasks.}}
\label{tab:sh}
\end{table*}

\begin{table*}[t]
\centering
\scriptsize
\setlength{\tabcolsep}{3pt}
\renewcommand{\arraystretch}{0.9}
\begin{tabular}{lcccccccccccc} 
\toprule
& Avg. & Avg. & Empty & Empty & Tray off & Tray in & Bottle in & Place & Remove  & Straighten & Slide \\
Models & Success (\%) $\uparrow$ & Rank $\downarrow$ & Container & Dishwasher & Oven & Oven & Fridge & 2 Cups & 2 Cups & Rope & \& Place \\
\midrule
RVT2+FT & 
4.1  & 3.7 & 
1.6 {\scriptsize± 0.4} & 0.0 {\scriptsize± 0.0} & 0.0 {\scriptsize± 0.0} & 1.2 {\scriptsize± 0.3} & 4.0 {\scriptsize± 0.7} & 1.3 {\scriptsize± 0.5} & 2.7 {\scriptsize± 0.8} & 2.7 {\scriptsize± 0.8} & 2.7 {\scriptsize± 0.7} \\
3DDA+FT & 
4.5 & 3.3 & 
1.6 {\scriptsize± 0.5} & 1.3 {\scriptsize± 0.6} & 1.6 {\scriptsize± 0.7} & 1.2 {\scriptsize± 0.5} & 4.3 {\scriptsize± 0.9} & 1.3 {\scriptsize± 0.5} & 5.3 {\scriptsize± 1.0} & 5.3 {\scriptsize± 1.1} & 0.0 {\scriptsize± 0.0} \\
ARP+FT & 
4.7 & 3.3 & 
1.3 {\scriptsize± 0.6} & 2.7 {\scriptsize± 0.9} & 5.3 {\scriptsize± 1.1} & 0.0 {\scriptsize± 0.0} & 2.7 {\scriptsize± 0.7} & 0.0 {\scriptsize± 0.0} & 4.0 {\scriptsize± 0.9} & 4.0 {\scriptsize± 0.8} & 5.3 {\scriptsize± 1.2} \\
\midrule
IMOP & 
7.4  & 2.6 & 
4.0 {\scriptsize± 0.8} & 1.3 {\scriptsize± 0.6} & 2.7 {\scriptsize± 0.9} & 5.3 {\scriptsize± 0.7} & 1.3 {\scriptsize± 0.5} & 2.7 {\scriptsize± 0.7} & 4.0 {\scriptsize± 0.8} & 4.0 {\scriptsize± 0.9} & 5.3 {\scriptsize± 1.0} \\
\rowcolor[HTML]{EFEFEF}
\textbf{ManiLong-Shot} & 
\textbf{30.2} \textcolor{black}{(\textbf{22.8\% $\uparrow$})} & 
\textbf{1.0} &
\textbf{28.0} {\scriptsize± 3.2} & 
\textbf{42.7} {\scriptsize± 2.8} & 
\textbf{37.3} {\scriptsize± 3.5} & 
\textbf{48.0} {\scriptsize± 2.1} & 
\textbf{18.7} {\scriptsize± 4.3} & 
\textbf{14.7} {\scriptsize± 3.9} & 
\textbf{24.0} {\scriptsize± 2.7} & 
\textbf{30.7} {\scriptsize± 2.5} & 
\textbf{26.7} {\scriptsize± 3.0} \\
\midrule
& Put & Take & Stack & Stack Cups & Put & Put  & Take  & Setup & Set  & Stack  & Block \\
Models & Item & Item & Cups & Blocks & 3 Books & Shoes & Shoes & Chess & Table & Blocks & Pyramid \\
\midrule
RVT2+FT & 
28.0 {\scriptsize± 1.9} & 29.3 {\scriptsize± 2.1} & 2.7 {\scriptsize± 0.6} & 0.0 {\scriptsize± 0.0} & 2.7 {\scriptsize± 0.7} & 0.0 {\scriptsize± 0.0} & 0.0 {\scriptsize± 0.0} & 0.0 {\scriptsize± 0.0} & 1.3 {\scriptsize± 0.5} & 0.0 {\scriptsize± 0.0} & 0.0 {\scriptsize± 0.0} \\
3DDA+FT & 
29.3 {\scriptsize± 2.0} & 30.7 {\scriptsize± 1.9} & 0.0 {\scriptsize± 0.0} & 0.0 {\scriptsize± 0.0} & 0.0 {\scriptsize± 0.0} & 1.3 {\scriptsize± 0.5} & 4.0 {\scriptsize± 0.8} & 1.3 {\scriptsize± 0.5} & 0.0 {\scriptsize± 0.0} & 0.0 {\scriptsize± 0.0} & 0.0 {\scriptsize± 0.0} \\
ARP+FT & 
18.7 {\scriptsize± 1.5} & 37.3 {\scriptsize± 2.4} & 8.0 {\scriptsize± 1.2} & 0.0 {\scriptsize± 0.0} & 0.0 {\scriptsize± 0.0} & 2.7 {\scriptsize± 0.6} & 0.0 {\scriptsize± 0.0} & 1.3 {\scriptsize± 0.5} & 0.0 {\scriptsize± 0.0} & 0.0 {\scriptsize± 0.0} & 0.0 {\scriptsize± 0.0} \\
\midrule
IMOP & 
40.0 {\scriptsize± 2.0} & 38.7 {\scriptsize± 2.1} & 4.0 {\scriptsize± 1.0} & 1.3 {\scriptsize± 0.6} & 9.3 {\scriptsize± 1.4} & 1.3 {\scriptsize± 0.5} & 1.3 {\scriptsize± 0.5} & 1.3 {\scriptsize± 0.5} & 2.7 {\scriptsize± 0.6} & 1.3 {\scriptsize± 0.5} & 1.3 {\scriptsize± 0.5} \\
\rowcolor[HTML]{EFEFEF}
\textbf{ManiLong-Shot} & 
\textbf{65.3} {\scriptsize± 2.4} & 
\textbf{76.0} {\scriptsize± 1.8} & 
\textbf{28.0} {\scriptsize± 3.1} & 
\textbf{18.7} {\scriptsize± 4.0} & 
\textbf{42.7} {\scriptsize± 2.7} & 
\textbf{29.3} {\scriptsize± 2.9} & 
\textbf{12.0} {\scriptsize± 4.6} & 
\textbf{9.3} {\scriptsize± 4.8} & 
\textbf{17.3} {\scriptsize± 3.5} & 
\textbf{8.0} {\scriptsize± 4.9} & 
\textbf{8.0} {\scriptsize± 4.9} \\
\bottomrule
\end{tabular}
\caption{\small{\textbf{OSIL Performance on 20 Unseen LH Tasks.} We report the mean and standard deviation of success rates over 5 random seeds for each task, along with the average success rate and average rank across all tasks. ``FT" means ``Fine-tuned with 5 demos".}}
\label{tab:lh}
\end{table*}

\begin{figure}[t!]
    \centering
    \includegraphics[width=\linewidth]{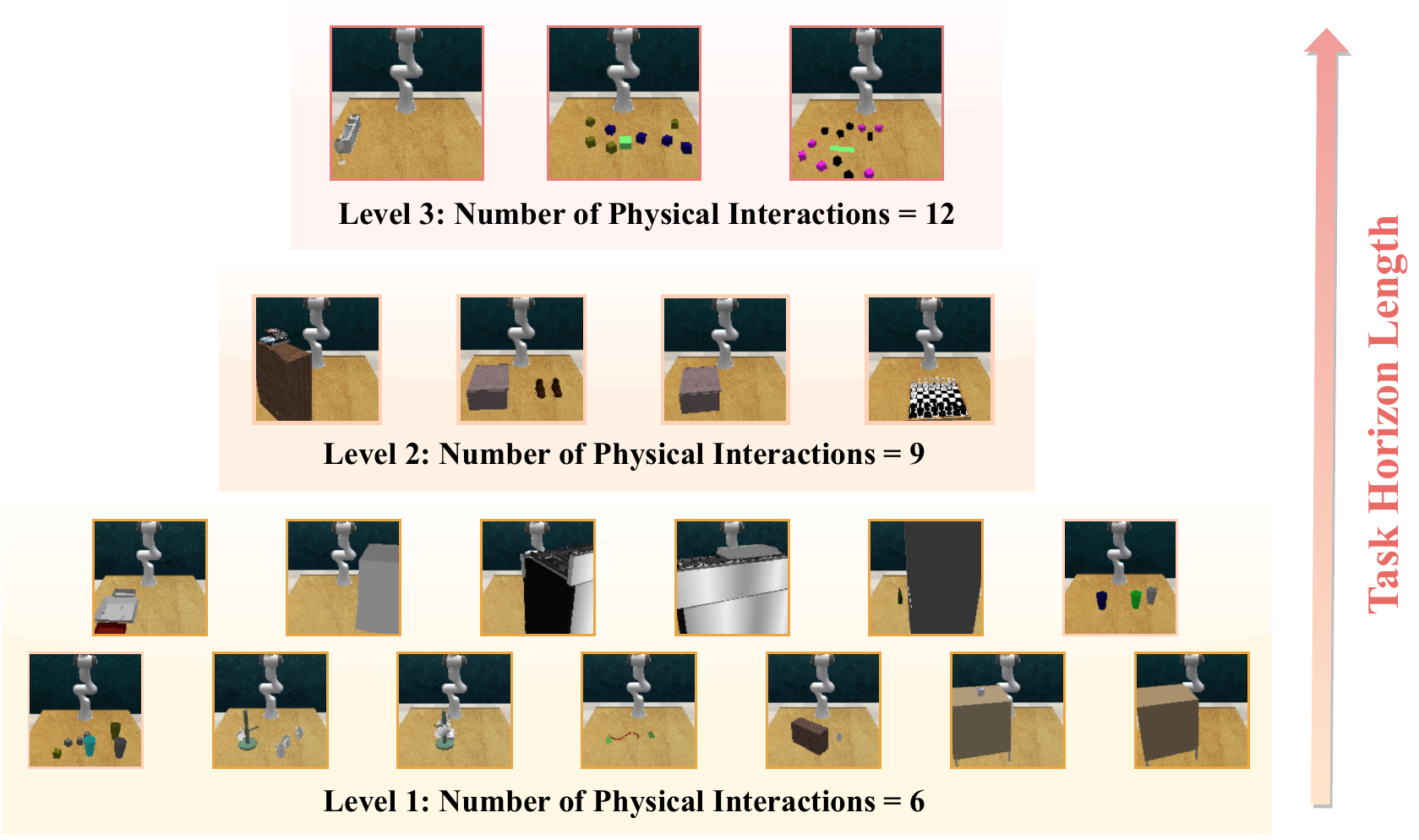}
    \caption{\small Visualization of 20 long-horizon manipulation tasks in \textbf{RLBench-Oneshot}, including 3 difficulty levels.}
    \label{fig:novel_tasks}
\end{figure}

\section{Experiments}
\label{sec:experiments}
We conduct extensive experiments in both simulated environments and real-world settings to address the following questions: 1) Can ManiLong-Shot achieve high success rates on SH tasks encountered during training, under different initial conditions?
2) Given a single successful demonstration, can ManiLong-Shot effectively one-shot generalize to unseen LH manipulation tasks without any fine-tuning?
3) Can ManiLong-Shot perform OSIL effectively in real-world LH tasks through sim-to-real transfer?
4) How does the heuristic design of the key modules affect ManiLong-Shot’s one-shot generalization performance on unseen LH tasks?

\subsection{Experimental Setup} 
\paragraph{Simulation Benchmark.}  
We select 30 tasks from the 100 tasks in the robotic manipulation simulation benchmark RLBench~\citep{james2020rlbench} to create the dedicated benchmark \textbf{RLBench-Oneshot} for evaluating the OSIL capabilities of robotic manipulation models. Based on the SH and LH task definitions in Sec.~\ref{formulation}, the benchmark includes 10 SH tasks and 20 LH tasks. The SH tasks involve a single round of \textit{pre-contact, grasping, and post-contact} interactions, with each task comprising 100 successful demonstration trajectories recorded from front, left/right shoulder, and wrist cameras using RGB-D observation.
The LH tasks are categorized into three difficulty levels based on the task horizon length, as shown in Fig.~\ref{fig:novel_tasks}: Level 1 includes 13 tasks with 6 physical interactions each; Level 2 includes 4 tasks with 9 interactions; Level 3 includes 3 tasks with 12 interactions. These levels comprehensively evaluate the model's OSIL ability on various LH tasks, with each LH task accompanied by one successful trajectory as a one-shot demonstration. More details on the RLBench-Oneshot benchmark tasks are provided in the Appendix~\ref{app:tasks}.

\paragraph{Real-World Setup.} We validate ManiLong-Shot with a UFactory xArm7 robot, utilizing RGB-D observations from RealSense D435 and D415 cameras. For motion planning, we integrate MoveIt~\citep{coleman2014reducing} to guide the arm to the end-effector poses. We set up three LH manipulation tasks in the real world: \textit{Stack Blocks}, \textit{Stack Cups}, and \textit{Place Cups}, each involving six physical interaction phases. For each task, we collect a successful trajectory through human teleoperation and conduct 5 one-shot generalization trials. In each trial, the object positions are slightly perturbed from those in the demonstration. This setup is used to evaluate the sim-to-real transfer of the ManiLong-Shot model pretrained in simulation, validating its effectiveness and robustness in real-world scenarios.

\paragraph{Baseline Models and Metrics.} 
We select ARP~\cite{zhang2025autoregressive}, 3DDA~\cite{ke20243d}, RVT2~\cite{goyal2024rvt}, and IMOP~\cite{zhang2024oneshot} as baseline models for comparison. The first three are state-of-the-art (SOTA) IL-based multi-task models on the RLBench benchmark, used to evaluate ManiLong-Shot’s performance on short-horizon (SH) tasks and its one-shot generalization on long-horizon (LH) tasks without fine-tuning. IMOP, the leading OSIL method for RLBench tasks, serves as a key baseline for assessing ManiLong-Shot's OSIL performance on unseen LH tasks.
Simulation evaluation metrics include average success rate and average rank. The success rate for each task is the mean and standard deviation from 25 trials with 5 random seeds, as RLBench-Oneshot generates different object layouts across trials. The average success rate is the mean across all tasks, while the average rank reflects overall performance by averaging rankings across tasks. For real-world robot experiments, we report the average success rate from 5 trials.

\subsection{Main Results}
\subsubsection{Performance Comparison on SH Tasks}
\label{experiments:base}
To evaluate ManiLong-Shot on training SH tasks, we assess policies learned solely from offline demonstrations, without any access to one-shot examples.  
Table~\ref{tab:sh} reports the performance of ManiLong-Shot compared to several baselines on 10 SH tasks from the RLBench-Oneshot benchmark, which are also used during training.  
In our main experiments, the interaction-aware task decomposition module is implemented using a rule-based approach. During evaluation, the interaction-aware region prediction network identifies distinct interaction phases by monitoring changes in the robot’s current state.  
Experimental results demonstrate that ManiLong-Shot achieves strong imitation learning performance, significantly outperforming three SOTA baselines. It reaches an average success rate of 90.4\%, yielding a \textbf{3.8\% improvement} over the best-performing baseline, 3DDA.  
ManiLong-Shot achieves the highest success rate on 6 out of 10 tasks, including precise manipulation (\textit{Open Drawer}, 100\% vs. ARP 92.8\%) and multi-step coordination (\textit{Put Money in Safe}, 100\% vs. 3DDA 97.6\%).  
These results highlight the effectiveness of ManiLong-Shot’s interaction-aware mechanism in maintaining consistent imitation learning performance across different stages of interaction.

\subsubsection{OSIL Performance on Novel LH Tasks}
\label{experiments:novel}

Table~\ref{tab:lh} shows the OSIL performance of ManiLong-Shot and baselines on 20 novel LH tasks from \textbf{RLBench-Oneshot}. RVT2+FT, 3DDA+FT, and ARP+FT denote models fine-tuned on each task using five successful demonstrations. Despite this few-shot adaptation, these models struggle to generalize, particularly on tasks involving novel scenes and interactions.
In contrast, for tasks that share scenes with training-time SH tasks (e.g., \textit{Put Item in Drawer} and \textit{Take Item out Drawer}, which share scenes with \textit{Open Drawer}), the fine-tuned models achieve moderate generalization. Notably, ManiLong-Shot reaches an average success rate of 30.2\% across all unseen LH tasks without task-specific fine-tuning, outperforming IMOP by \textbf{22.8\%}.
On the most challenging unseen LH tasks, ManiLong-Shot achieves significant results without fine-tuning, such as \textit{Set Table} (17.4\% vs. IMOP 0.0\%), \textit{Stack Blocks} (8.0\% vs. IMOP 0.0\%), and \textit{Block Pyramid} (8.0\% vs. IMOP 0.0\%). Notably, \textit{Stack Blocks} and \textit{Block Pyramid} involve long interaction sequences and require precise manipulation. These results underscore the effectiveness of ManiLong-Shot’s three core modules. 
For complex LH tasks, ManiLong-Shot decomposes tasks into interaction-aware primitives aligned with physical interaction phases, enhancing robustness to variations in object poses and dynamics. By performing region prediction and matching at each phase, the system improves execution stability and overall success.

\begin{figure}[t!]
\centering
\includegraphics[width=0.9\linewidth]{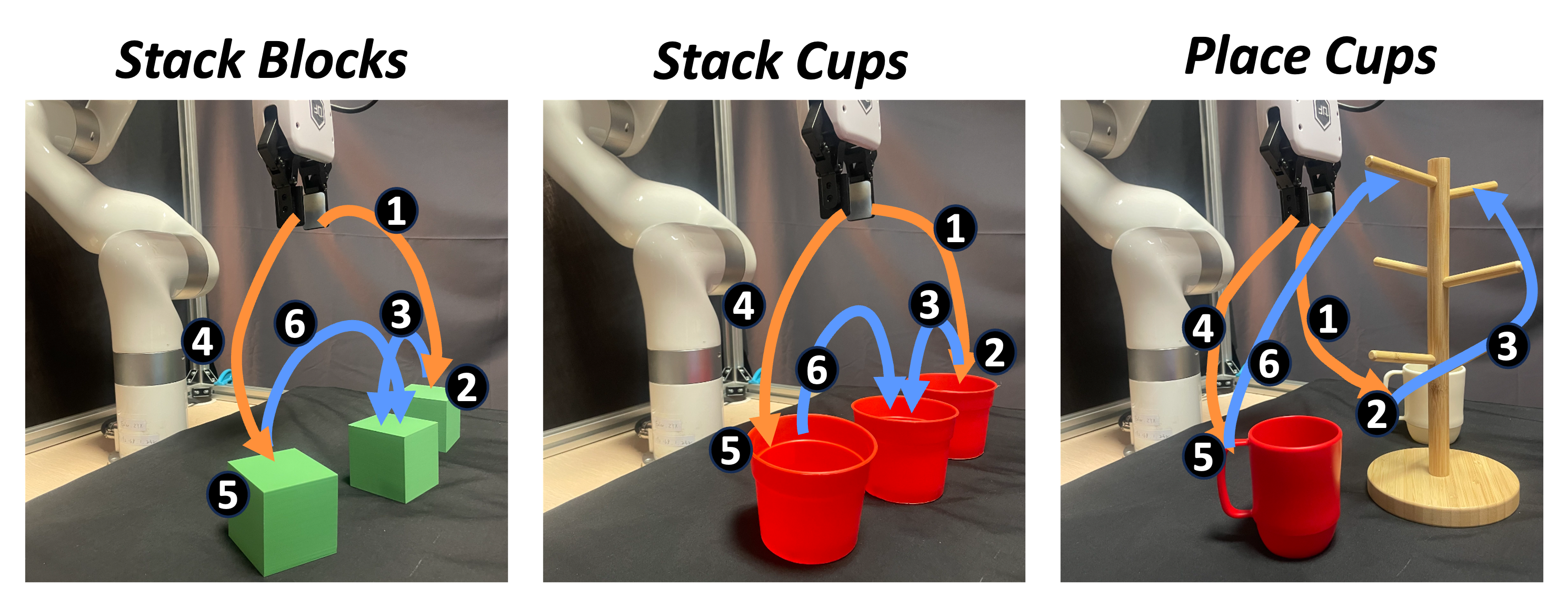}
\caption{\small Three real-world LH tasks with their physical interaction visualizations.}
\label{fig:real-robot-exp}
\end{figure}

\begin{table}[t!]
\centering
\small
\setlength{\tabcolsep}{5pt}
\renewcommand{\arraystretch}{0.9}
\begin{tabular}{lcc}\toprule
\multirow{2}{*}{Task} &\multicolumn{2}{c}{Success Rate} \\\cmidrule{2-3}
&IMOP & \textbf{ManiLong-Shot} \\\midrule
Stack Blocks & 60\% & \textbf{80\%} \\
Stack Cups & 20\% & \textbf{60\%} \\
Place Cups & 20\% & \textbf{40\%} \\\midrule
Average & 33.3\% & \textbf{60.0\% \textcolor{black}{(26.7\% $\uparrow$)}}  \\
\bottomrule
\end{tabular}
\caption{\small \textbf{The OSIL Sim-to-Real Transfer Performance.}}
\label{tab:real-robot-exp}
\end{table}

\paragraph{Real-world Experiments}
We evaluate ManiLong-Shot’s OSIL performance in real-world LH manipulation tasks via sim-to-real transfer on three representative tasks: \textit{Stack Blocks}, \textit{Stack Cups}, and \textit{Place Cups} (as illustrated in Fig.~\ref{fig:real-robot-exp}). For each task, a single successful demonstration trajectory is collected through human teleoperation of the robot arm, with the data collection process visualized in the figure. Table~\ref{tab:real-robot-exp} reports the average success rates of ManiLong-Shot and IMOP over five OSIL trials on each task, with each evaluation performed under slightly varied object layouts. The results show that ManiLong-Shot achieves effective OSIL performance on real-world LH manipulation tasks, outperforming IMOP with an average success rate improvement of approximately 26.7\%.

\subsection{Ablation Studies}
\begin{figure}[t!]
\includegraphics[width=\linewidth]{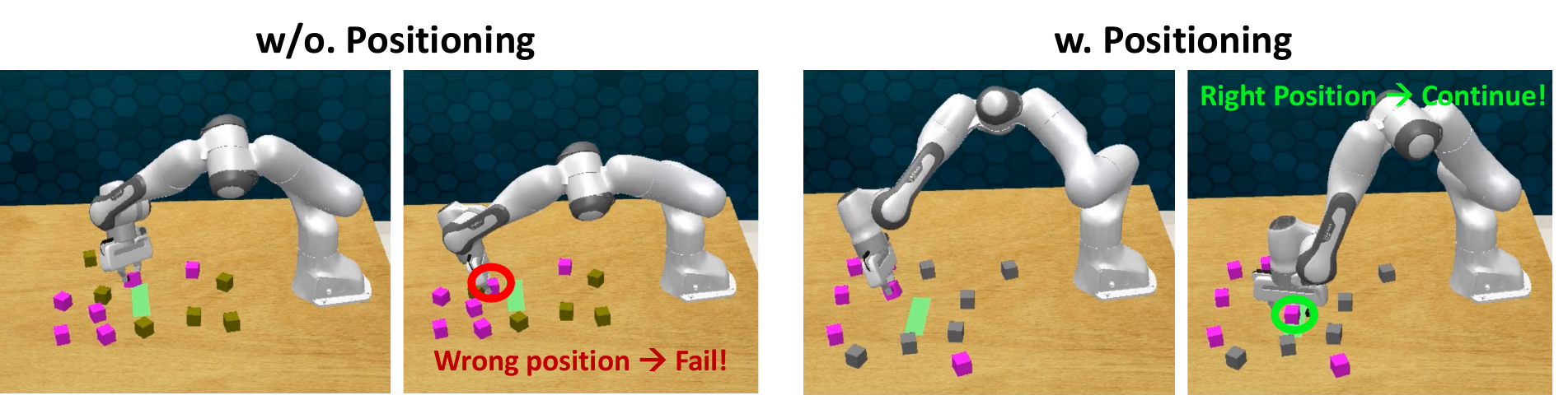} 
    \centering
    \includegraphics[width=\linewidth]{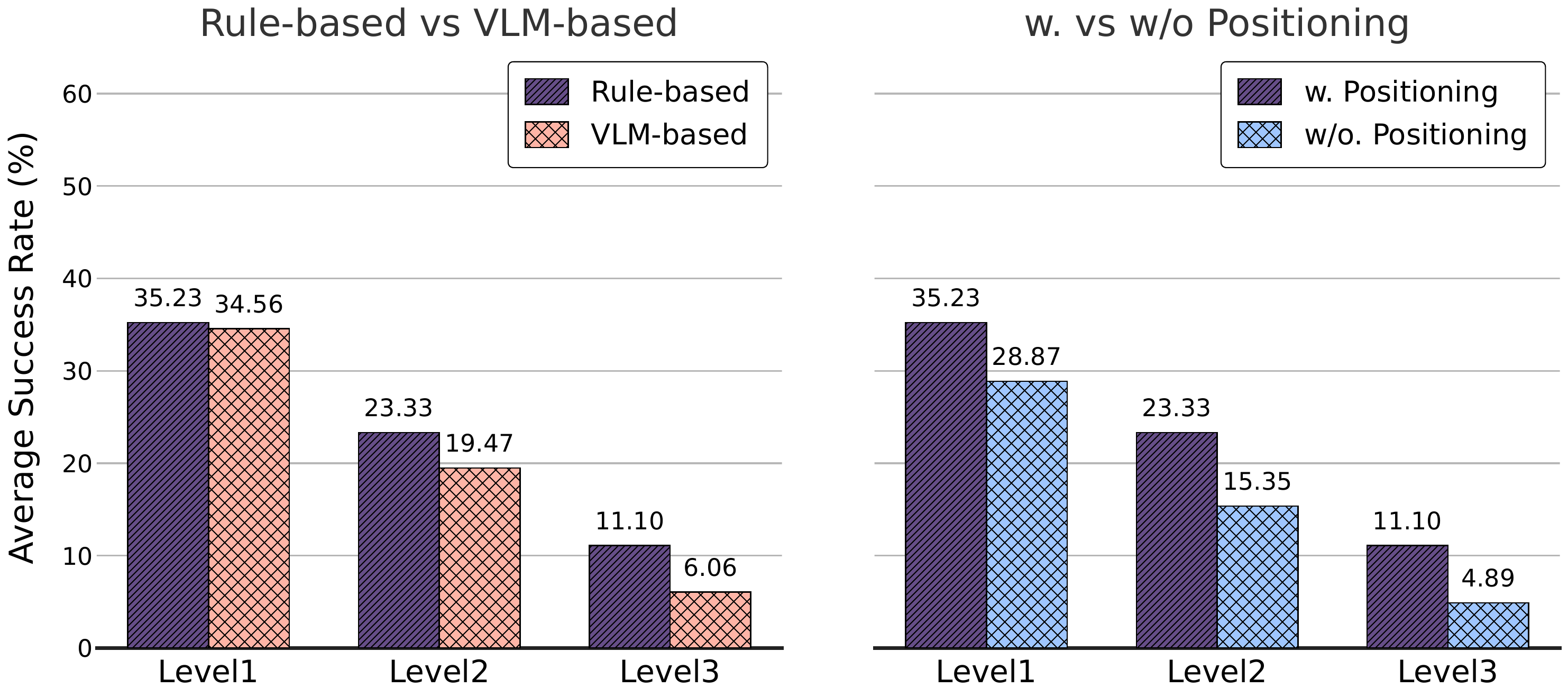}
    \caption{\small \textbf{Ablation Study of Key Modules in ManiLong-Shot.} Comparison of average success rates on the LH tasks across three difficulty levels in RLBench-Oneshot.}
    \label{fig:ablation}
\end{figure}

\paragraph{VLM-based vs. Rule-based.} 
We investigate how different task decomposition strategies within ManiLong-Shot’s interaction-aware task decomposition module affect OSIL performance on unseen LH tasks. As shown in Fig.~\ref{fig:ablation}(left), the VLM-based decomposition consistently underperforms the rule-based approach across all difficulty levels, primarily due to the instability of VLM reasoning. As task difficulty increases, leading to more primitives to identify and execute, the performance gap widens. Nevertheless, ManiLong-Shot with VLM-based decomposition still significantly outperforms all baseline models on unseen LH tasks, validating the framework's effectiveness in one-shot generalization. Detailed prompts for VLM are presented in Appendix~\ref{app:vlm}.

\paragraph{W. Positioning vs. w/o. Positioning.} We further investigate the effect of the positioning network in ManiLong-Shot’s interaction-aware invariant region prediction module, particularly for invariant region prediction during the post-contact phase. As shown in Fig.~\ref{fig:ablation}(right), across all three difficulty levels, removing the positioning network in this phase may result in inaccurate placement of the grasped object. Such misalignment often disrupts the execution of subsequent sub-task primitives, which ultimately impedes the successful completion of the overall LH task and causes a notable decline in the system’s OSIL performance on unseen LH tasks. Fig.~\ref{fig:ablation}(top) visualizes the differences in ManiLong-Shot’s performance on the \textit{Block Pyramid} task when placing the first block in the target region, comparing the cases with and without the positioning network.

\section{Conclusion and Limitations}
We present ManiLong-Shot, a novel OSIL framework tailored for long-horizon prehensile manipulation tasks. ManiLong-Shot leverages the physical interaction between the robot and the object to decompose tasks into \textit{pre-contact}, \textit{grasping}, and \textit{post-contact} primitives, predicting interaction-aware invariant regions for each phase. Using these predictions, ManiLong-Shot performs region matching and pose prediction through correspondences during the one-shot learning phase. We evaluate the performance of ManiLong-Shot through extensive experiments on both simulated and real-world robots, demonstrating its effective and robust OSIL ability for long-horizon manipulation.

\paragraph{Limitations and Future Work.} 
A primary limitation of ManiLong-Shot lies in its decomposition strategy, which defines manipulation in terms of three discrete phases grounded in physical contact. While effective for many long-horizon prehensile tasks, this limits applicability to tasks with extended post-contact tool use, such as wiping or pouring, where interactions are continuous and hard to decompose. Future work will focus on extending the framework to handle more general long-horizon tasks, cross-embodiment scenarios, and complex, temporally continuous behaviors.

\section*{Acknowledgements}
This work is supported in part by National Natural Science Foundation of China (62192783, 62276128), Jiangsu Natural Science Foundation (BK20243051), Jiangsu Science and Technology Major Project (BG2024031), the Fundamental Research Funds for the Central Universities(14380128, KG202514), the Collaborative Innovation Center of Novel Software Technology and Industrialization, and the Program of China Scholarship Council (Grant No. 202306190157).

\bibliography{aaai2026}

\clearpage

\appendix
\newpage
\section*{Technical Appendix}
\addcontentsline{toc}{section}{Appendix}

In this technical appendix, we provide more information on the ManiLong-Shot framework. First, in Sec.~\ref{app:training_theo}, we offer additional insights into the assumptions and network architecture involved in ManiLong-Shot. Then, in Sec~\ref{app:training_archi}, we provide a detailed explanation of the structure of the State Routing Network used in the Interaction-aware Invariant Region Matching network. Sec.~\ref{app:training_details} provides the training details and objective for ManiLong-Shot.
Next, in Sec.~\ref{app:eval}, we present an evaluation flowchart for ManiLong-Shot’s one-shot imitation learning (OSIL) on previously unseen long-sequence tasks, followed by a more detailed description of this process. Additionally, we include additional experimental details in Sec.~\ref{app:experiments}, where we first give an in-depth introduction to the tasks involved in the RLBench-Oneshot simulation benchmark (Sec.~\ref{app:tasks}) and further elaborate on the computational resources and relevant codebase used in the experimental design (Sec.~\ref{app:experiments_setup}).
Finally, we provide detailed prompts for VLM-based task decomposition in Sec.~\ref{app:vlm}, including the specific prompt template.

\appsection{Further Training Details of ManiLong-Shot}
\label{app:training}

\appsubsection{More Theoretical Details}
\label{app:training_theo}

\paragraph{Definition of Invariant Regions.}  
ManiLong-Shot aims to generalize the action policy from a single successful demonstration in a long-horizon task $\mathcal{T}_k \in \mathcal{T}^{\text{lh}}$. Given this task, the observed state transitions $(\hat{s}_i, \hat{a}_i, \hat{s}_{i+1})$ from the demonstration (where $\tau^{\text{lh}} = \{(\hat{s}_t, \hat{a}_t, \hat{s}_{t+1})\}_{t=0}^{H}$) and new real-time states $s_j$ encountered during execution, the goal is to generalize the operation strategy.
We assume that $\hat{s}_i$ and $s_j$ should share the same optimal operation action. That is, despite different scene configurations, they should apply the same operation strategy. Based on this assumption, ManiLong-Shot can map the action pose $\mathbf{T}_i$ from state $\hat{s}_i$ to a new action pose $\mathbf{T}_j$ to complete the task at state $s_j$\footnote{Symbol conventions are consistent with the main text, where $\hat{s}_i$ and $\hat{a}_i$ represent states and actions from the test task, and $\mathbf{T}_i$ represents the action pose transformation.}.

Specifically, given a task $\mathcal{T}_k \in \mathcal{T}^{\text{lh}}$, the invariant region of state $\hat{s}_i$ is defined as the set of 3D points in its point cloud that remain stable across all states $s_j$ sharing the same optimal operation strategy. These invariant regions allow the robot to generalize from the demonstrated state to the real-time state, mapping action pose $\mathbf{T}_i$ to $\mathbf{T}_j$ and performing the task.
According to the definition proposed by IMOP~\cite{zhang2024oneshot}, the invariant region of $\hat{s}_i$ in task $\mathcal{T}_k$ is defined as:

\begin{multline}
\mathcal{I}(\hat{s}_i \mid \mathcal{T}_k) = \left\{ p \in \hat{s}_i \mid 
\forall s_j \in \mathcal{S},\, \hat{s}_i \equiv s_j,\,
\exists q \in s_j : \right. \\
\left. \left\| \mathbf{T}_i^{-1} p - \mathbf{T}_j^{-1} q \right\| < \epsilon \right\},
\label{app:invariant}
\end{multline}
where $(\mathbf{T}_i, \lambda_i) = \pi^*(\hat{s}_i \mid \mathcal{T}_k)$, and $(\mathbf{T}_j, \lambda_j) = \pi^*(s_j \mid \mathcal{T}_k)$.

In Definition~\ref{app:invariant}, $\hat{s}_i$ represents a demonstrated state in the point cloud of a single demonstration trajectory in the long-horizon task $\mathcal{T}_k$, and $\hat{s}_i \equiv s_j$ to indicate that $s_i$ and $s_j$ share the same optimal manipulation action. $\mathcal{I}(\hat{s}_i \mid \mathcal{T}_k)$ is its invariant region, a set of 3D points in the $\mathbf{T}_i$ coordinate system that remain stable across all states $s_j$ sharing the optimal operation. $\mathbf{T}_i$ is the end-effector pose when performing the optimal action at state $\hat{s}_i$.

\paragraph{Definition of Correspondence Matrix.} The invariant region $\mathcal{I}(\hat{s}_i)$ is predicted as the set of activated points through a point-wise sigmoid over $\hat{s}_i$. Then, we follow the procoess in IMOP~\cite{zhang2024oneshot}, applying graph cross-attention layers between the KNN graph of $\mathcal{I}(\hat{s}_i)$ and the real-time state $s_j$ to extract the point-wise features $h_{\mathcal{I}(\hat{s}_i)} \in \mathbb{R}^{|\mathcal{I}(s_i)| \times D}$ and $h_{s_j} \in \mathbb{R}^{|s_j| \times D}$, where $D$ denotes the size of the feature dimension. Finally, we perform a dual softmax matching~\citep{li2022lepard} between $h_{\mathcal{I}(s_i)}$ and $h_{s_j}$ to obtain the correspondence matrix $\mathbf{C} \in \{0, 1\}^{|\mathcal{I}(s_i)| \times |s_j|}$:
$\mathbf{C} = \operatorname{softmax}(h_{\mathcal{I}(\hat{s}_i)} \cdot h_{s_j}^\top) \cdot \operatorname{softmax}(h_{s_j} \cdot h_{\mathcal{I}(\hat{s}_i)}^\top)^\top,$
where $\operatorname{softmax}$ is applied to each row. The correspondence matrix $C$ maps each point in the invariant region $\mathcal{I}(\hat{s}_i)$ to the real-time state $s_j$. The matched points in $s_j$ constitute $\mathcal{I}(s_j)$.

\paragraph{Correspondence-based Pose Regression.} 
As described in Sec.~\ref{formulation} in the main paper, we directly adopt the correspondence-based pose regression algorithm~\citep{zhang2024oneshot} to estimate the robot’s action pose $\mathbf{T}_j$ at real-time state $s_j$. The core objective of this algorithm is to solve an optimization problem that minimizes the transformation between a reference pose $\mathbf{T}_i$ and the target pose $\mathbf{T}_j$, thereby inferring the robot's pose in $s_j$.
The optimization objective is defined as:
$\mathbf{T}_j = \arg\min_{\mathbf{T} \in \mathrm{SE}(3)} 
\left\| \mathbf{T} \, \mathbf{T}_i^{-1} \, P_{\mathcal{I}(\hat{s}_i)} \, \mathbf{C} - P_{{s}_j} \right\|,$
where $\mathbf{T}_i$ is the demonstrated action pose at state $s_i$, representing the robot's specific configuration (including rotation and translation). $P_{\mathcal{I}(\hat{s}_i)}$ and $P_{s_j}$ denote the point clouds of the invariant region in state $\hat{s}_i$ (i.e., the parts of the local environment that remain static), and the complete scene in state $\hat{s}_j$, respectively.
In this optimization problem, $\mathbf{T}$ is the transformation matrix to be solved, which belongs to the special Euclidean group $\mathrm{SE}(3)$, representing transformations that include both rotation and translation in 3D space. By solving for $\mathbf{T}$, the objective is to minimize the discrepancy between the transformed point cloud $\mathbf{T} \mathbf{T}_i^{-1} P_{\mathcal{I}(\hat{s}_i)}$ and the target scene $P_{s_j}$.
Specifically, $P_{\mathcal{I}(\hat{s}_i)}$ is a set of 3D points defined by the invariant region in state $\hat{s}_i$, representing the spatial distribution of fixed objects or surfaces, while $P_{s_j}$ is the full scene point cloud at state $s_j$. The matrix $\mathbf{C}$ is a correspondence matrix that encodes the best matching relationship between each point in $P_{\mathcal{I}(\hat{s}_i)}$ and $P_{s_j}$.
The norm $\left\| \cdot \right\|$ in the objective function computes the Euclidean distance between the transformed point cloud and the target point cloud. By selecting an appropriate transformation matrix $\mathbf{T}$, we aim to align the transformed source points as closely as possible with the target scene, thus determining a pose consistent with the target configuration.

From a physical perspective, this process can be interpreted as estimating the robot's new pose $\mathbf{T}_j$ in state $s_j$ by adjusting the previous pose $\mathbf{T}_i$ from state $\hat{s}_i$, leveraging the geometric properties of the invariant region. By exploiting prior motion patterns and current environmental constraints, the robot can accurately predict the new action pose.
The optimal pose $\mathbf{T}_j$ corresponds to the solution of the above objective function when the point correspondences in $\mathbf{C}$ yield the minimum overall pairwise distance after applying the transformation $\mathbf{T}_j \mathbf{T}_i^{-1}$. Therefore, the pose regression problem can be addressed by learning visual features that align with this objective function. Following the computation method in IMOP, we adopt the differentiable Procrustes operator from Lepard~\citep{li2022lepard} to solve the least-squares problem.

\begin{figure}[h]
    \centering
    \includegraphics[width=\linewidth]{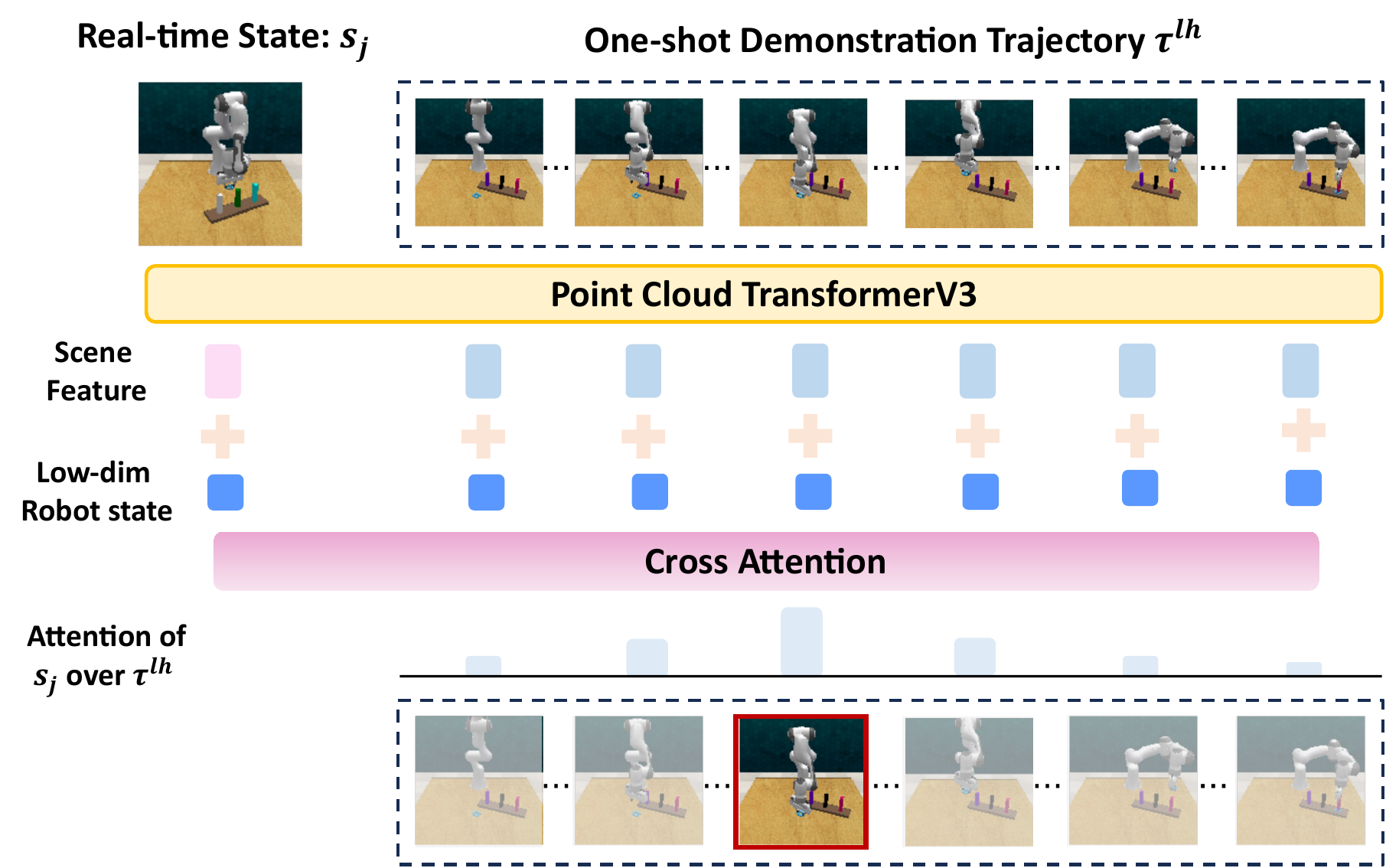}
    \caption{The Architecture of the State Routing Network, adopted from IMOP.}
    \label{fig:state-routing}
\end{figure}

\begin{figure}[H]
    \centering
    \includegraphics[width=\linewidth]{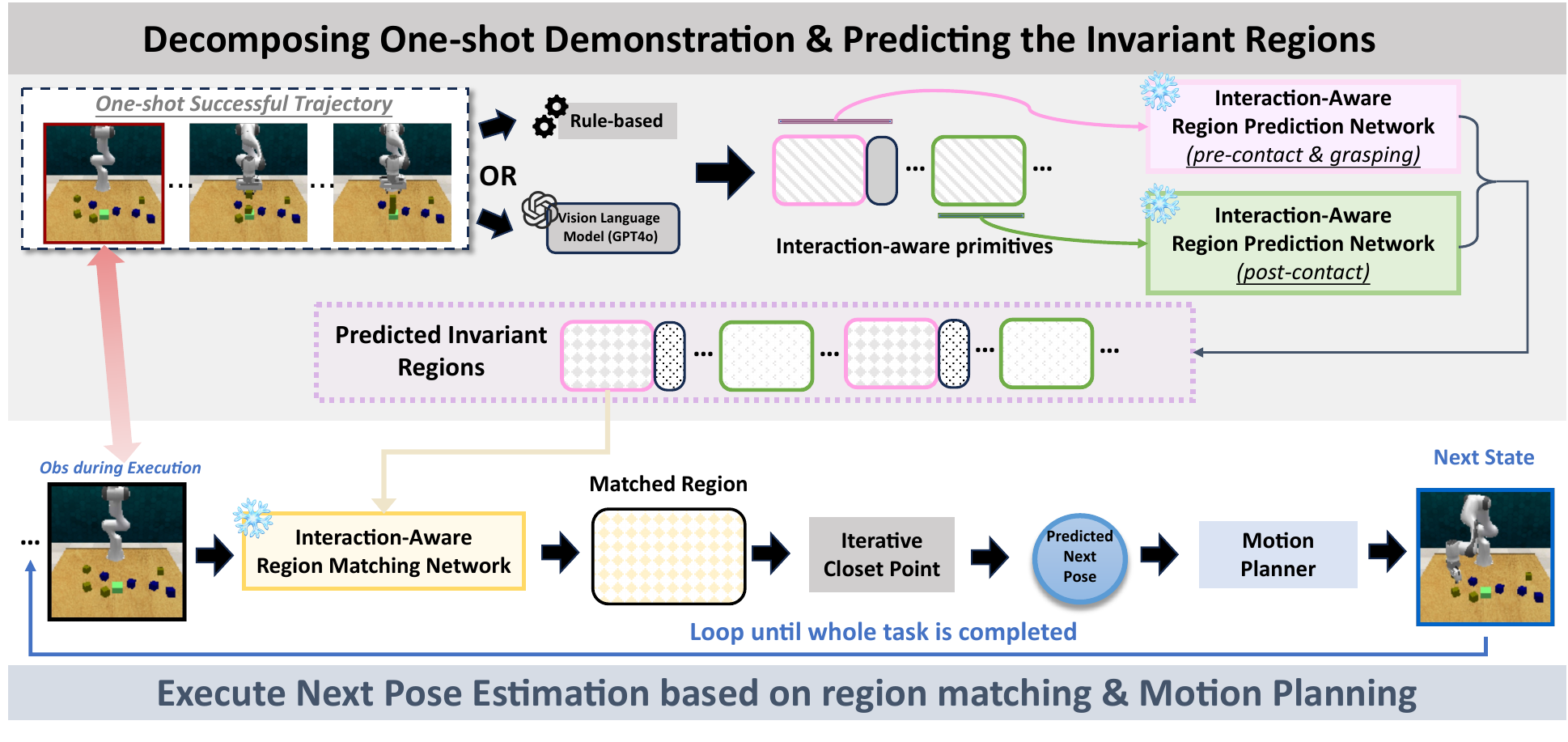}
    \caption{The evaluation pipeline of ManiLong-Shot.}
    \label{fig:eval_pipeline}
\end{figure}

\appsubsection{More Architecture Details}
\label{app:training_archi}

We mainly refer to the State Routing Network proposed by \citeauthor{zhang2024oneshot} (\citeyear{zhang2024oneshot}), which is used to select the demonstration frame $\hat{s}_i$ from a single successful demonstration trajectory $\tau^{\mathrm{lh}}$ based on the current real state $s_j$. Unlike the network proposed by Zhang et al., we first use the PTv3 backbone network (where PTv2 is used in IMOP) to extract the scene-level features for both the real state $s_j$ and each state in the demonstration trajectory $\tau^{\mathrm{lh}}$. Then, we concatenate the scene-level features with the low-dimensional internal robot states, including joint positions and time steps. Next, we apply cross-attention to the features of multiple states. The state with the strongest attention to the real state $s_j$ is selected as the demonstration state $\hat{s}_i$. The specific state routing network architecture is shown in Fig.~\ref{fig:state-routing}.

\appsubsection{More Training Details}
\label{app:training_details}
The training procedure of ManiLong-Shot follows that of IMOP~\cite{zhang2024oneshot}, assuming access to instance segmentation results during training. We utilize the segmentation masks provided by RLBench~\citep{james2020rlbench}, but notably, ManiLong-Shot does not require segmentation information during inference.
Unlike IMOP, which generates macro-steps by compressing each state using the keyframe extraction method from C2F-ARM~\citep{james2022coarse}, ManiLong-Shot adopts a different approach to temporal abstraction. Specifically, for each decomposed interaction phase, it samples one frame every 5 frames (inclusive of the start and end frames) to define macro-steps. If the interval between the final frame and the previous macro-step is less than 5 frames, this constraint is relaxed, and the final frame is directly included as an independent macro-step.

At each training iteration, a pair of trajectories, $\tau_i$ and $\tau_j$, is sampled such that $|\tau^{sh}_i| = |\tau^{sh}_j|$. The state routing network is trained using focal loss~\citep{lin2017focal} to predict that $\overline{s}_i \equiv \overline{s}_j$ (i.e., $\overline{s}_i$ and $\overline{s}_j$ share the same manipulation action) when they occur at the same timestep.
For training the interaction-aware invariant region matching network, let $\{c_{i,n}\}_{n=1}^{N}$ represent the $N$ instance segments in $\overline{s}_i$, where each $c_{i,n}$ is a set of 3D points corresponding to an object instance. We estimate the ground-truth matched region $\mathcal{I}(\overline{s}_i)$ as follows:
\begin{align}
    \mathcal{I}(\overline{s}_i) = \arg \min_{c_{i, n}} \mathbb{E}_{\substack{p \in c_{i, n} \\ \overline{s}_i \equiv \overline{s}_j}} \left[ \min_{\substack{q \in c_{j, m} \\ \operatorname{cls}(c_{i,n}) = \operatorname{cls}(c_{j,m})}} \| T_i^{-1} p - T_j^{-1} q \| \right],
\end{align}
where $\operatorname{cls}(c_{i,n})$ denotes the object class of the segment $c_{i,n}$.
The ground-truth $\mathcal{I}(\overline{s}_i)$ corresponds to the instance segment in $\overline{s}_i$ that has the minimal displacement in the frame of the action pose $T$. The point-level correspondence between $\overline{s}_i$ and $\overline{s}_j$ is defined as the set:
\begin{align}
    \left\{(p, q) \in \mathcal{I}(\overline{s}_i) \times \mathcal{I}(\overline{s}_j) \mid q = \arg\min_{q \in \mathcal{I}(\overline{s}_j)} \| T_i^{-1} p - T_j^{-1} q \| \right\},
\end{align}
where the point-wise displacement is minimized in the action pose frame.

\appsection{Further Evaluation Details of ManiLong-Shot}
\label{app:eval}

The evaluation pipeline of ManiLong-Shot for unseen long-horizon manipulation tasks is presented in Fig.~\ref{fig:eval_pipeline}.

During evaluation, ManiLong-Shot takes as input a single successful demonstration of the task and follows a structured pipeline to execute the entire manipulation sequence on a novel scene.
The process begins with the Interaction-Aware Task Decomposition Module, which analyzes the demonstration trajectory and segments it into a sequence of interaction-aware primitives. Each primitive represents a robot-object physically meaningful interaction phase, i.e., \textit{pre-contact, grasping, and post-contact}. This segmentation is guided by changes of robot state, enabling the system to reason over temporally localized behaviors.
For each segmented interaction phase, the Interaction-Aware Region Prediction Network predicts an invariant region in the environment that is critical for that phase's execution. These invariant regions are defined as spatial areas in the scene that are essential for achieving the corresponding subgoal and are expected to remain consistent across different task instances. The prediction leverages both the point cloud representation of the environment and internal robot state information, such as joint configurations and timestamps.

Once the task has been decomposed and the invariant regions have been predicted, the framework proceeds to the execution stage. In this stage, ManiLong-Shot continuously receives real-time observations from the robot, such as RGB-D images and proprioceptive inputs. At each timestep, it employs the Interaction-Aware Region Matching Network to identify which predicted invariant region best matches the current observation. This is achieved by computing a correspondence matrix between the predicted region and the observed scene, effectively aligning the demonstration with the current task instance.
Based on the matching results, the framework performs pose prediction by estimating the optimal transformation that maps the demonstration pose to the current context. This pose is then used by the motion planning module to generate an executable action for the robot. The execution is carried out in a closed-loop fashion: after executing each motion primitive, the system updates its observation inputs, reassesses the matching with the predicted invariant regions, and proceeds to the next primitive as determined by the state routing network.
This loop continues iteratively until the robot completing the full task. Thanks to its modular design and reliance on interaction-centric cues, ManiLong-Shot is capable of robustly adapting to novel task layouts, object configurations, and environmental conditions, all from a single successful demonstration.

\appsection{Additional Experimental Details}
\label{app:experiments}

\appsubsection{Details of RLBench-Oneshot}
\label{app:tasks}

\begin{figure}
    \centering
    \includegraphics[width=\linewidth]{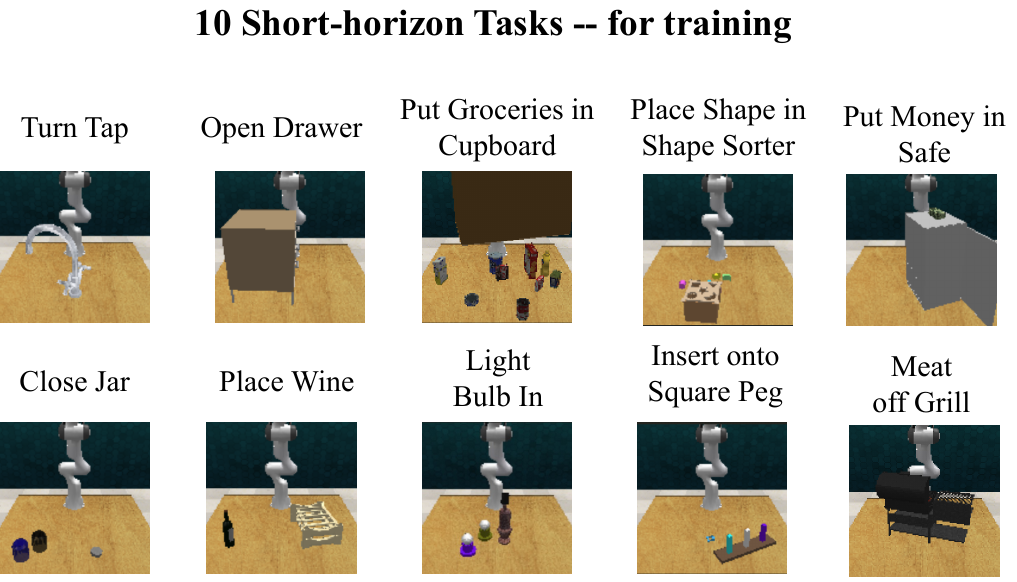}
    \caption{Visulization of 10 Short-horizon Manipulation Tasks in RLBench-Oneshot.}
    \label{fig:short_tasks}
\end{figure}

\paragraph{Short-horizon Tasks.} We first introduce the 10 short-horizon manipulation tasks in RLBench-Oneshot, as shown in Fig.~\ref{fig:short_tasks}.

\vspace{1em}  

\noindent\textbf{(1) Close Jar}

\noindent\textbf{Task:} Close the jar by placing the lid on the jar.\\
\noindent\textbf{Object:} 1 block and 2 colored jars.\\
\noindent\textbf{Success Metric:} The jar lid is successfully placed on the jar as detected by the proximity sensor. \\
\noindent\textbf{Number of Physical Interactions (\textit{Pre-contact, Grasping, Post-contact}):} 3

\vspace{1em}  

\noindent\textbf{(2) Open Drawer}

\noindent\textbf{Task:} Open the drawer by gripping the handle and pulling it open.\\
\noindent\textbf{Object:} 1 drawer.\\
\noindent\textbf{Success Metric:} The drawer is successfully opened to the desired position as detected by the joint condition on the drawer's joint. \\
\noindent\textbf{Number of Physical Interactions (\textit{Pre-contact, Grasping, Post-contact}):} 3

\vspace{1em}  

\noindent\textbf{(3) Light Bulb In}

\noindent\textbf{Task:} Screw in the light bulb by picking it up from the holder and placing it into the lamp.\\
\noindent\textbf{Object:} 2 light bulbs, 2 holders, and 1 lamp stand.\\
\noindent\textbf{Success Metric:} The light bulb is successfully screwed into the lamp and detected by the proximity sensor. \\
\noindent\textbf{Number of Physical Interactions (\textit{Pre-contact, Grasping, Post-contact}):} 3

\vspace{1em}  

\noindent\textbf{(4) Meat Off Grill}

\noindent\textbf{Task:} Take the specified meat off the grill and place it next to the grill.\\
\noindent\textbf{Object:} 1 piece of chicken, 1 piece of steak, and 1 grill.\\
\noindent\textbf{Success Metric:} The specified meat is successfully removed from the grill and detected by the proximity sensor. \\
\noindent\textbf{Number of Physical Interactions (\textit{Pre-contact, Grasping, Post-contact}):} 3

\vspace{1em}

\noindent\textbf{(5) Insert onto Square Peg}

\noindent\textbf{Task:} Insert a square ring onto the spoke with the specified color.  \\
\noindent\textbf{Object:} 1 square, and 1 spoke platform with three color spokes.\\
\noindent\textbf{Success Metric:} The square ring is successfully placed onto the correctly colored spoke. \\
\noindent\textbf{Number of Physical Interactions (\textit{Pre-contact, Grasping, Post-contact}):} 3

\vspace{1em}

\noindent\textbf{(6) Turn Tap}

\noindent\textbf{Task:} Turn either the left or right handle of the tap. Left and right are defined with respect to the
faucet orientation.  \\
\noindent\textbf{Object:} 1 faucet with 2 handles.\\
\noindent\textbf{Success Metric:} The revolute joint of the specified handle is at least $90^o$ off from the starting
position. \\
\noindent\textbf{Number of Physical Interactions (\textit{Pre-contact, Grasping, Post-contact}):} 3

\vspace{1em}

\noindent\textbf{(7) Put Groceries in Cupboard}

\noindent\textbf{Task:} Grab the specified object and put it in the cupboard above. The scene always contains 9 YCB
objects that are randomly placed on the tabletop. \\
\noindent\textbf{Object:} 9 YCB objects, and 1 cupboard (that hovers in the air like magic).\\
\noindent\textbf{Success Metric:} The specified object is inside the cupboard. \\
\noindent\textbf{Number of Physical Interactions (\textit{Pre-contact, Grasping, Post-contact}):} 3

\vspace{1em}

\noindent\textbf{(8) Place Shape in Shape Sorter}

\noindent\textbf{Task:} Pick up the specified shape and place it inside the correct hole in the sorter. There are always 4 distractor shapes, and 1 correct shape in the scene.  \\
\noindent\textbf{Object:} 5 shapes, and 1 sorter. \\
\noindent\textbf{Success Metric:} The square ring is successfully placed onto the correctly colored spoke. \\
\noindent\textbf{Number of Physical Interactions (\textit{Pre-contact, Grasping, Post-contact}):} 3

\vspace{1em}

\noindent\textbf{(9) Put Money in Safe}

\noindent\textbf{Task:} Pick up the stack of money and put it inside the safe on the specified shelf. The shelf has
three placement locations: top, middle, bottom. \\
\noindent\textbf{Object:} 1 stack of money, and 1 safe. \\
\noindent\textbf{Success Metric:} The square ring is successfully placed onto the correctly colored spoke. \\
\noindent\textbf{Number of Physical Interactions (\textit{Pre-contact, Grasping, Post-contact}):} 3

\vspace{1em}

\noindent\textbf{(10) Place Wine }

\noindent\textbf{Task:} Grab the wine bottle and put it on the wooden rack at one of the three specified locations:
left, middle, right. The locations are defined with respect to the orientation of the wooden rack.  \\\noindent\textbf{Object:} 1 wine bottle, and 1 wooden rack. \\
\noindent\textbf{Success Metric:} The square ring is successfully placed onto the correctly colored spoke. \\
\noindent\textbf{Number of Physical Interactions (\textit{Pre-contact, Grasping, Post-contact}):} 3




\begin{figure}
    \centering
    \includegraphics[width=\linewidth]{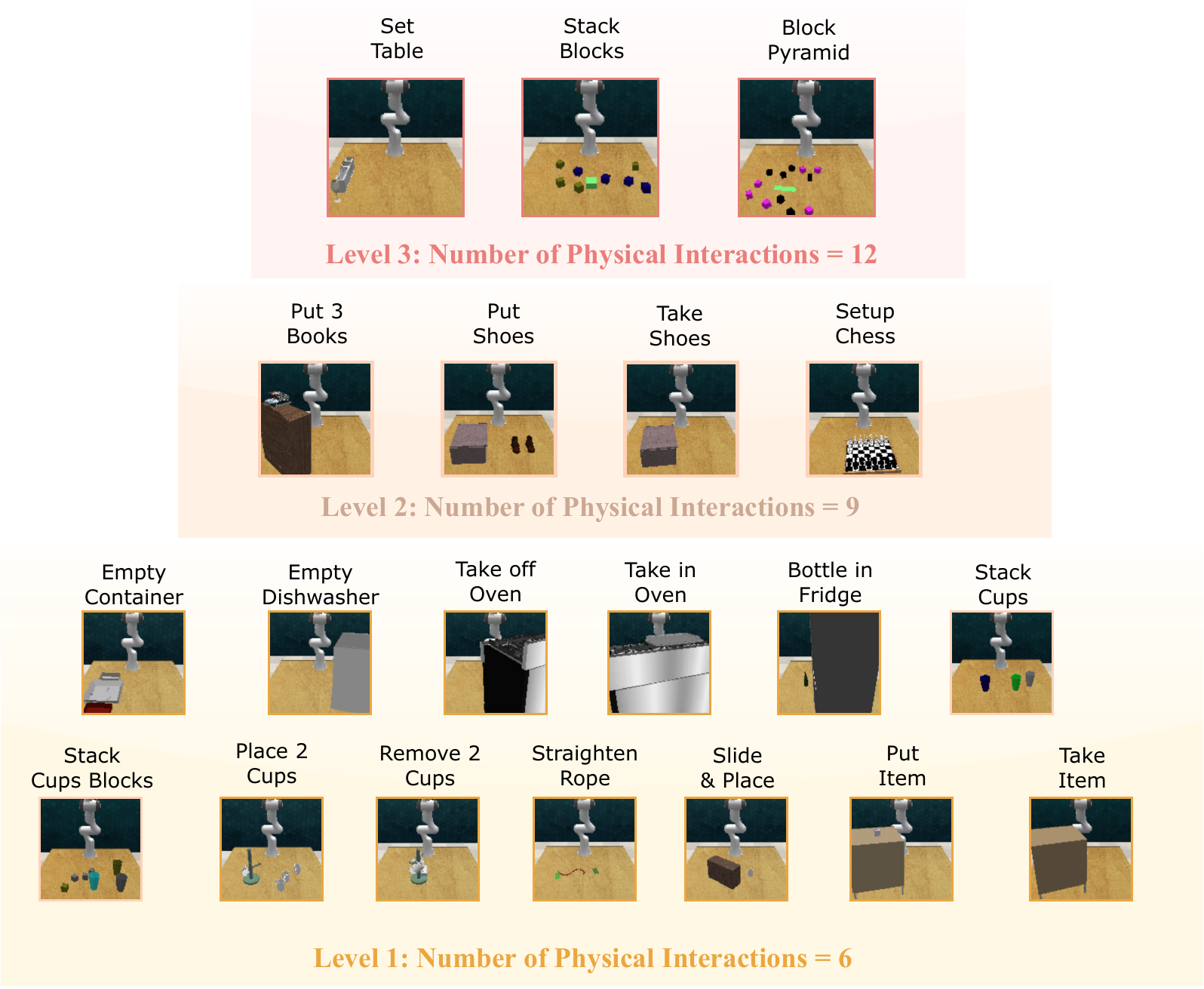}
    \caption{Visulization of 20 Long-horizon Manipulation Tasks in RLBench-Oneshot.}
    \label{fig:long_tasks}
\end{figure}

\paragraph{Long-horizon Tasks.} Next, We detail the 20 long-horizon manipulation tasks in RLBench-Oneshot, as shown in Fig.~\ref{fig:long_tasks}.

\vspace{1em}  

\noindent\textbf{(1) Empty Container}

\noindent\textbf{Task:} Remove all procedurally generated objects from the large central container and place them into the target small container of a specified color.\

\noindent\textbf{Object:} 1 large container, 2 small containers, 2 procedural objects.\

\noindent\textbf{Success Metric:} All procedural objects are detected inside the correct small container by a proximity sensor.\

\noindent\textbf{Number of Physical Interactions (\textit{Pre-contact, Grasping, Post-contact}):} 6

\vspace{1em}  

\noindent\textbf{(2) Empty Dishwasher}

\noindent\textbf{Task:} Open the dishwasher, slide the rack out, and remove the plate from inside.\\
\noindent\textbf{Object:} 1 dishwasher, 1 dishwasher plate.\\
\noindent\textbf{Success Metric:} The plate is no longer detected inside the dishwasher by a proximity sensor (i.e., it has been removed from the dishwasher).\\
\noindent\textbf{Number of Physical Interactions (\textit{Pre-contact, Grasping, Post-contact}):} 6

\vspace{1em}  

\noindent\textbf{(3) Take off Oven}

\noindent\textbf{Task:} Open the oven, grasp the tray, and remove it from the oven.\\
\noindent\textbf{Object:} 1 tray, 1 oven.\\
\noindent\textbf{Success Metric:} The tray is no longer detected inside the oven by a proximity sensor, and the robot's gripper is not holding anything.\\
\noindent\textbf{Number of Physical Interactions (\textit{Pre-contact, Grasping, Post-contact}):} 6

\vspace{1em}  

\noindent\textbf{(4) Take in Oven}

\noindent\textbf{Task:} Open the oven door, pick up the tray, and place it inside the oven.\\
\noindent\textbf{Object:} 1 tray, 1 oven.\\
\noindent\textbf{Success Metric:} The tray is detected inside the oven by a proximity sensor, and the robot's gripper is empty.\\
\noindent\textbf{Number of Physical Interactions (\textit{Pre-contact, Grasping, Post-contact}):} 6

\vspace{1em}

\noindent\textbf{(5) Bottle in Fridge}

\noindent\textbf{Task:} Open the fridge door, pick up the bottle, and place it inside the fridge.\\
\noindent\textbf{Object:} 1 bottle, 1 fridge.\\
\noindent\textbf{Success Metric:} The bottle is successfully placed inside the fridge as detected by a proximity sensor, and the robot's gripper is empty.\\
\noindent\textbf{Number of Physical Interactions (\textit{Pre-contact, Grasping, Post-contact}):} 6

\vspace{1em}

\noindent\textbf{(6) Stack Cups}

\noindent\textbf{Task:} Stack the other two cups on top of the designated cup based on its color.\\
\noindent\textbf{Object:} 3 cups.\\
\noindent\textbf{Success Metric:} The designated cup is detected inside the stack, and the robot's gripper is empty.\\
\noindent\textbf{Number of Physical Interactions (\textit{Pre-contact, Grasping, Post-contact}):} 6

\vspace{1em}

\noindent\textbf{(7) Stack Cups Blocks}

\noindent\textbf{Task:} Stack the other two cups with the same color of a specific block on top of the rest cup.\\
\noindent\textbf{Object:} 3 cups.\\
\noindent\textbf{Success Metric:} The designated cup is detected inside the stack, and the robot's gripper is empty.\\
\noindent\textbf{Number of Physical Interactions (\textit{Pre-contact, Grasping, Post-contact}):} 6

\vspace{1em}

\noindent\textbf{(8) Place 2 Cups}

\noindent\textbf{Task:} Place 2 cups on the cup holder by sliding their handles onto the spokes.\\
\noindent\textbf{Object:} 3 cups, 3 spokes on the cup holder.\\
\noindent\textbf{Success Metric:} The 2 cups are successfully placed on the cup holder and detected by proximity sensors, and the robot's gripper is empty.\\
\noindent\textbf{Number of Physical Interactions (\textit{Pre-contact, Grasping, Post-contact}):} 6.

\vspace{1em}

\noindent\textbf{(9) Remove 2 Cups}

\noindent\textbf{Task:} Remove 2 cups from the cup holder and place them on the table.\\
\noindent\textbf{Object:} 3 cups, 1 cup holder.\\
\noindent\textbf{Success Metric:} The specified number of cups is successfully removed from the holder and detected by the proximity sensors, and the robot's gripper is empty.\\
\noindent\textbf{Number of Physical Interactions (\textit{Pre-contact, Grasping, Post-contact}):} 6

\vspace{1em}

\noindent\textbf{(10) Straighten Rope}

\noindent\textbf{Task:} Straighten the rope by pulling each end until it is taut.\\
\noindent\textbf{Object:} 1 rope (with identifiable head and tail).\\
\noindent\textbf{Success Metric:} Both ends of the rope are detected as straightened by proximity sensors at the head and tail, indicating the rope is taut on the table.\\
\noindent\textbf{Number of Physical Interactions (\textit{Pre-contact, Grasping, Post-contact}):} 6

\vspace{1em}

\noindent\textbf{(11) Slide Cabinet Open and Place Cups}

\noindent\textbf{Task:} Open the cabinet door and place a cup inside.\\
\noindent\textbf{Object:} 1 cup, cabinet with two sides (left and right).\\
\noindent\textbf{Success Metric:} The cup is successfully placed inside the cabinet, detected by a proximity sensor, and the robot's gripper is empty.\\
\noindent\textbf{Number of Physical Interactions (\textit{Pre-contact, Grasping, Post-contact}):} 6

\vspace{1em}

\noindent\textbf{(12) Put Item in Drawer}

\noindent\textbf{Task:} Place an item inside one of the specified drawers.\\
\noindent\textbf{Object:} 1 item, 3 drawers (bottom, middle, top).\\
\noindent\textbf{Success Metric:} The item is successfully placed in the specified drawer, detected by a proximity sensor, and the robot's gripper is empty.\\
\noindent\textbf{Number of Physical Interactions (\textit{Pre-contact, Grasping, Post-contact}):} 6

\vspace{1em}

\noindent\textbf{(13) Take Item Out of Drawer}

\noindent\textbf{Task:} Take an item out of one of the specified drawers.\\
\noindent\textbf{Object:} 1 item, 3 drawers (bottom, middle, top).\\
\noindent\textbf{Success Metric:} The item is successfully taken out of the specified drawer, detected by a proximity sensor, and the robot's gripper is empty.\\
\noindent\textbf{Number of Physical Interactions (\textit{Pre-contact, Grasping, Post-contact}):} 6

\vspace{1em}

\noindent\textbf{(14) Put 3 Books on Bookshelf}

\noindent\textbf{Task:} Place 3 books onto the bookshelf.\\
\noindent\textbf{Object:} 3 books.\\
\noindent\textbf{Success Metric:} The specified number of books are successfully placed on the bookshelf and detected by a proximity sensor, and the robot's gripper is empty.\\
\noindent\textbf{Number of Physical Interactions (\textit{Pre-contact, Grasping, Post-contact}):} 9

\vspace{1em}

\noindent\textbf{(15) Put Shoes in Box}

\noindent\textbf{Task:} Place shoes inside a box.\\
\noindent\textbf{Object:} 2 shoes, 1 box.\\
\noindent\textbf{Success Metric:} Both shoes are successfully placed inside the box, detected by a proximity sensor, and the robot's gripper is empty.\\
\noindent\textbf{Number of Physical Interactions (\textit{Pre-contact, Grasping, Post-contact}):} 9

\vspace{1em}

\noindent\textbf{(16) Take Shoes Out of Box}

\noindent\textbf{Task:} Remove shoes from a box and place them on a table.\\
\noindent\textbf{Object:} 2 shoes, 1 box.\\
\noindent\textbf{Success Metric:} Both shoes are successfully taken out of the box and placed on the table, detected by a proximity sensor, and the robot's gripper is empty.\\
\noindent\textbf{Number of Physical Interactions (\textit{Pre-contact, Grasping, Post-contact}):} 9

\vspace{1em}

\noindent\textbf{(17) Setup Chess}

\noindent\textbf{Task:} Arrange 3 chess pieces on a chessboard in their initial positions.\\
\noindent\textbf{Object:} Multiple chess pieces (king, queen, bishops, knights, rooks, pawns) and a chessboard.\\
\noindent\textbf{Success Metric:} The specified number of chess pieces are successfully placed in their initial positions on the chessboard, detected by proximity sensors, and the robot's gripper is empty.\\
\noindent\textbf{Number of Physical Interactions (\textit{Pre-contact, Grasping, Post-contact}):} 9

\vspace{1em}

\noindent\textbf{(18) Set The Table}

\noindent\textbf{Task:} Arrange dishes and cutlery on a table in preparation for a meal.\\
\noindent\textbf{Object:} 1 plate, 1 fork, 1 knife, 1 spoon, 1 glass.\\
\noindent\textbf{Success Metric:} The plate, fork, knife, spoon, and glass are successfully placed on the table, detected by proximity sensors, and the robot's gripper is empty.\\
\noindent\textbf{Number of Physical Interactions (\textit{Pre-contact, Grasping, Post-contact}):} 12

\vspace{1em}

\noindent\textbf{(19) Stack Blocks}

\noindent\textbf{Task:} Stack 4 blocks on top of each other.\\
\noindent\textbf{Object:} Up to 4 target blocks and 4 distractor blocks.\\
\noindent\textbf{Success Metric:} The specified number of target blocks are successfully stacked, detected by a proximity sensor, and the robot's gripper is empty.\\
\noindent\textbf{Number of Physical Interactions (\textit{Pre-contact, Grasping, Post-contact}):} 12

\vspace{1em}

\noindent\textbf{(20) Block Pyramid}

\noindent\textbf{Task:} Stack blocks to form a pyramid.\\
\noindent\textbf{Object:} 6 target blocks and 6 distractor blocks.\\
\noindent\textbf{Success Metric:} Three target blocks are successfully stacked to form a pyramid shape, detected by proximity sensors, and the robot's gripper is empty.\\
\noindent\textbf{Number of Physical Interactions (\textit{Pre-contact, Grasping, Post-contact}):} 12

\appsubsection{More Details of Experimental Setup}
\label{app:experiments_setup}
For all simulation experiments, we use a single NVIDIA GeForce RTX 4090 GPU with 24GB of VRAM for both training and testing. The experimental framework is based on the open-source IMOP codebase available at \url{https://github.com/mlzxy/imop/tree/main}, all network configurations and hyperparameters follow those used in IMOP. the dataset construction is derived from the task code of RLBench at \url{https://github.com/stepjam/RLBench/tree/master/rlbench/tasks}.

\newpage

\appsection{Prompt Template for VLM-base Task Decomposition}
\label{app:vlm}

\begin{mybox}
\footnotesize

\textbf{Task Objective:} \\
Please review a time-ordered sequence of data and identify the key phases or sub-tasks within it. The task should be segmented into the following repeating phases: `pre-contact`, `grasping`, `post-contact`. Continue identifying and segmenting the phases in this sequence (pre-contact → grasping → post-contact → pre-contact → grasping → post-contact...) until the entire task is completed.

\vspace{0.8em}
\textbf{Input Format Explanation:} \\
You will receive a sequence of data with multiple time steps. Each time step will contain the following fields (which can be omitted or replaced as per the actual task):

\begin{lstlisting}
[
  {
    "t": 0,
    "obs": "observation at timestep 0",
    "action": "action taken at timestep 0",
    "event": "optional event log",
    "sensor": {"gripper_state": ..., "joint_velocity": ..., ...}
  },
  ...
]
\end{lstlisting}

\vspace{0.5em}
\textbf{Output Format Requirements:} \\
Please return the identified phases in JSON format. Each phase should contain the following fields:

\begin{lstlisting}
[
  {
    "stage": "name of the stage (e.g., pre-contact, grasping, post-contact)",
    "start": start_timestep,
    "end": end_timestep,
    "reason": "brief reason for identifying this segment"
  },
  ...
]
\end{lstlisting}

\textit{Please ensure:}
\begin{itemize}
  \item All phases are continuous and non-overlapping.
  \item Each phase boundary is determined based on input data (e.g., changes in gripper state, joint velocity, action intent, or state transitions).
  \item The sequence of phases must repeat in the following order until the entire task is completed: `pre-contact → grasping → post-contact → pre-contact → grasping → post-contact ...`
\end{itemize}

\vspace{0.5em}
\textbf{Optional Task Instructions (choose based on needs):}
\begin{itemize}
  \item \textbf{[Task Type A]: Physical Interaction Phase Identification} \\
  Segment physical interaction phases based on signals like object contact, grasping, placing, etc.

  \item \textbf{[Task Type B]: Language Instruction Task Phase Analysis} \\
  Segment the sequence of sub-tasks based on robot actions performed as per language instructions (e.g., ``open drawer and place cup'').

  \item \textbf{[Task Type C]: Anomaly Detection \& Phase Re-labeling} \\
  Detect any failed or interrupted phases and re-label the task phases accordingly.

  \item \textbf{[Task Type D]: Multi-Modal Cooperative Understanding} \\
  Identify structured semantic phases based on image + action/sensor data for task progress.
\end{itemize}

\vspace{0.5em}
\textbf{Example Output (for reference):}
\begin{lstlisting}
[
  {
    "stage": "pre-contact",
    "start": 0,
    "end": 14,
    "reason": "The end-effector moves toward the target object while the gripper remains fully open. Although joint velocities may temporarily drop to zero during motion, the gripper posture stays unchanged and does not indicate preparation for grasping. The phase ends when the robot completes positioning near the object and the joint velocity converges to zero, marking a transition to a static pre-grasp state."
  },
  {
    "stage": "grasping",
    "start": 15,
    "end": 17,
    "reason": "The gripper begins to close, and a noticeable change in joint velocity suggests initiation of a grasping action."
  },
  {
    "stage": "post-contact",
    "start": 18,
    "end": 40,
    "reason": "The object has been grasped and is being moved toward the target location."
  },
  {
    "stage": "pre-contact",
    "start": 41,
    "end": 45,
    "reason": "The robot moves the grasped object toward the next task location with the gripper still in place, maintaining a consistent joint velocity."
  },
  {
    "stage": "grasping",
    "start": 46,
    "end": 48,
    "reason": "The gripper adjusts to secure the object more firmly as the robot approaches the final target."
  },
  {
    "stage": "post-contact",
    "start": 49,
    "end": 60,
    "reason": "The object is now placed at the target location and the gripper releases it."
  }
]
\end{lstlisting}

\end{mybox}

\end{document}